\documentclass{article}

\usepackage{PRIMEarxiv}

\usepackage[utf8]{inputenc}
\usepackage{amsmath, amsthm}
\usepackage{amssymb}
\usepackage{hyperref}
\usepackage{url}
\usepackage{tcolorbox}
\usepackage{subcaption}
\usepackage{booktabs}
\usepackage{enumitem}
\usepackage{tabularx}
\usepackage{amsmath,amsfonts,bm}

\def\eqref#1{equation~\ref{#1}}
\def\1{\bm{1}}

\DeclareMathAlphabet{\mathsfit}{\encodingdefault}{\sfdefault}{m}{sl}
\SetMathAlphabet{\mathsfit}{bold}{\encodingdefault}{\sfdefault}{bx}{n}

\usepackage{graphicx}
\usepackage{natbib}
\usepackage{tikz}
\usetikzlibrary{
  positioning,   % for "above=… of …"
  arrows.meta,   % for the new arrow styles ("<->")
  shapes,        % if you want more node shapes later
  matrix,        % for the legend matrix
  calc           % for coordinate calculations
}
\usepackage{thmtools} % both already present in earlier code
\theoremstyle{plain}
\newtheorem{theorem}{Theorem}[section]

\newtheorem{lemma}[theorem]{Lemma}
\newtheorem{corollary}[theorem]{Corollary}
\theoremstyle{definition}

\newtheorem{assumption}[theorem]{Assumption}
\theoremstyle{remark}
\newtheorem{remark}[theorem]{Remark}

\title{Effective Sample Size and Generalization Bounds for Temporal Networks}

\author{
  Barak Gahtan\thanks{Corresponding author: \texttt{barakgahtan@cs.technion.ac.il}} \\
  Department of Computer Science\\
  Technion -- Israel Institute of Technology\\
  Haifa, Israel\\
  \And
  Alex M. Bronstein \\
  Department of Computer Science\\
  Technion -- Israel Institute of Technology\\
  Haifa, Israel\\
  \textit{and}\\
  ISTA -- Institute of Science and Technology\\
  Vienna, Austria\\
}

\begin{document}
\maketitle

\begin{abstract}
Learning from time series is fundamentally different from learning from i.i.d.\ data:
temporal dependence can make long sequences effectively information-poor, yet standard
evaluation protocols conflate sequence length with statistical information. We propose
a dependence-aware evaluation methodology that controls for effective sample size
$N_{\text{eff}}$ rather than raw length $N$, and provide end-to-end generalization
guarantees for Temporal Convolutional Networks (TCNs) on $\beta$-mixing sequences.
Our analysis combines a blocking/coupling reduction that extracts $B = \Theta(N/\log N)$
approximately independent anchors with an architecture-aware Rademacher bound for
$\ell_{2,1}$-norm-controlled convolutional networks, yielding $O(\sqrt{D\log p / B})$
complexity scaling in depth $D$ and kernel size $p$. Empirically, we find that stronger
temporal dependence can \emph{reduce} generalization gaps when comparisons control for
$N_{\text{eff}}$ - a conclusion that reverses under standard fixed-$N$ evaluation, with
observed rates of $N_{\text{eff}}^{-0.9}$ to $N_{\text{eff}}^{-1.2}$ substantially faster
than the worst-case $O(N^{-1/2})$ mixing-based prediction. Our results suggest that
dependence-aware evaluation should become standard practice in temporal deep learning
benchmarks.
\end{abstract}


\section{Introduction}
Modern deep architectures, notably Temporal Convolutional Networks (TCNs) \cite{lea2017temporal,bai2018empirical} and Transformer variants \cite{vaswani2017attention}, underpin state-of-the-art forecasting and representation learning across domains ranging from clinical monitoring to large-scale operational forecasting and management \cite{lim2021temporal,oreshkin2019n}. Despite this success, two fundamental gaps limit our understanding of temporal deep learning.

Gap 1: Evaluation on dependent data is confounded. A common practice is to compare models by varying the raw sequence length $N$ or by holding $N$ fixed while changing dependence strength (e.g., correlation). However, for dependent sequences, $N$ is a poor proxy for the amount of statistical information: strong temporal correlation can drastically reduce the number of effectively independent observations (``effective sample size'')\cite{geyer1992practical,sokal1997monte}. As a result, ``standard'' comparisons at equal $N$ conflate two distinct effects: (1) changes in temporal structure (dependence) and (2) changes in information content. This confounding can systematically bias conclusions about if dependence helps or hinders learning.

Gap 2: Architectural scaling under dependence lacks clear guarantees. Classical generalization analyses rely on independence and therefore do not directly apply to time series. While mixing-based learning theory \cite{yu1994rates,kuznetsov2014generalization} provides tools to analyze dependence, it often does not expose how modern architectural choices (depth, kernel size, norm control) affect sample complexity in deep temporal models. In contrast, norm-based i.i.d. analyses yield explicit architectural dependence (e.g., $\sqrt{D}$ rather than exponential in depth) under explicit norm control \cite{neyshabur2015norm,golowich2018size}. A central challenge is to retain such architecture-aware scaling while handling temporal dependence.

Our approach: effective-sample-size matching with supporting theory. We address these gaps with a methodology-first approach. On the empirical side, we introduce evaluation protocols that control for an \textbf{effective sample size} $N_{\text{eff}}$, i.e., a proxy for the number of ``nearly independent'' learning-relevant observations contained in a length-$N$ dependent sequence. Because $N_{\text{eff}}$ is not uniquely defined in general, we adopt a definition that is \emph{aligned with our theory}: under $\beta$-mixing, the key control quantity in our bounds is the \emph{anchor count} induced by blocking (denoted $B$), and we instantiate $N_{\text{eff}}$ so that comparisons equalize this effective information budget. This enables comparisons across dependence regimes by separating changes in information content from changes in temporal structure. On the theoretical side, we combine a blocking/coupling reduction for $\beta$-mixing sequences with an i.i.d.\ architecture-dependent complexity bound for norm-controlled convolutional networks (via $\ell_{2,1}$ filter-group norm constraints). The resulting bounds are conservative but provide a baseline: they establish learnability under dependence and make explicit how architectural scaling laws interact with effective information.

We make three contributions:
\begin{enumerate}
    \item \textbf{Fair-comparison methodology for dependent sequences.} We propose to match $N_{\text{eff}}$ rather than raw $N$ when the goal is to compare models or dependence regimes on equal information budgets.
    \item \textbf{Empirical findings enabled by fair comparison.} Applying this methodology to synthetic autoregressive processes and physiological sequences reveals regimes in which stronger dependence is associated with smaller generalization gaps at fixed $N_{\text{eff}}$, a phenomenon that is obscured (and can appear reversed) under fixed-$N$ evaluation.
    \item \textbf{Architecture-aware generalization baseline under $\beta$-mixing.} We provide end-to-end bounds for TCNs on exponentially $\beta$-mixing sequences, achieving explicit dependence on depth (via a $\sqrt{D}$ factor) and mild polylogarithmic dependence on kernel size. Under exponential $\beta$-mixing, the dependent-to-i.i.d.\ reduction yields an effective anchor sample size $B=\Theta(N/\log N)$, inducing an additional $\sqrt{\log N}$ factor relative to the i.i.d.\ $1/\sqrt{N}$ rate.
\end{enumerate}

We distinguish between \textbf{standard evaluation} (comparisons at fixed raw length $N$) and \textbf{fair comparison} (comparisons that control for effective sample size $N_{\text{eff}}$). Section~\ref{sec:related} reviews related work on dependent-data learning and norm-based complexity control. Section~\ref{sec:prem} provides preliminaries. Section~\ref{sec:methods} presents our dependence-aware generalization baseline, and Section~\ref{sec:experiments} reports empirical results. We conclude in Section~\ref{sec:conclusion}; the appendix contains full proofs and additional experimental details.

\section{Related Work}\label{sec:related}
\textbf{Generalization under dependence.} Classical PAC-style generalization theory is typically developed for i.i.d.\ samples, while time series violate this assumption. A long line of work studies concentration and uniform convergence for stationary mixing processes, often via blocking/coupling arguments that reduce dependent sequences to collections of approximately independent blocks (e.g., early empirical-process rates for $\beta$-mixing sequences~\cite{yu1994rates}, nonparametric time-series prediction through adaptive model selection~\cite{meir2000nonparametric}, and surveys of mixing tools~\cite{bradley2005basic}). Building on these ideas, Mohri and Rostamizadeh~\cite{mohri2008rademacher} developed Rademacher-complexity bounds for $\beta$-mixing sequences, and later stability-based bounds were also derived for mixing processes~\cite{mohri2009stability}. Alternative dependent-learning viewpoints include discrepancy-based generalization analyses~\cite{kuznetsov2014generalization}, as well as PAC-Bayes approaches for weakly dependent sequences (e.g.,~\cite{alquier2018simpler}). Recent work by Ab\'el\`es et al.~\cite{abeles2024generalization} proposes an online-to-PAC framework with delayed feedback to control dependence, which is complementary to our focus here. We focus on absolute regularity ($\beta$-mixing) because it supports total-variation coupling, the key tool behind our anchor-based dependent-to-i.i.d.\ reduction. Alternative online-learning approaches (e.g., sequential Rademacher complexity~\cite{rakhlin2010online}) handle adversarial sequences but express guarantees in regret terms rather than sample complexity.

\textbf{Norm-based generalization for deep networks (i.i.d.).} Classical VC-dimension analyses provide architecture-dependent bounds that scale with parameter counts~\cite{bartlett1998almost,bartlett2019nearly}, but these can be loose for modern over-parameterized networks. Norm control has become a standard way to obtain architecture-dependent generalization guarantees for deep networks. For feedforward networks, seminal bounds scale with products of layer norms and improve depth dependence relative to VC-style analyses (e.g.,~\cite{neyshabur2015norm,bartlett2017spectrally,golowich2018size}). These results motivate using norm budgets as a proxy for function-class capacity and help explain why deep networks can generalize despite over-parameterization. Our theoretical ingredient follows this tradition: we rely on a norm-controlled complexity bound for convolutional layers and then lift it to dependent data through a blocking/coupling reduction.

\textbf{Convolutional networks and weight sharing.}
Generalization analyses for CNNs must explicitly account for parameter sharing and the structured linear operators induced by convolution. Long and Sedghi~\cite{long2019generalization} provide generalization bounds for deep CNNs that are independent of the input resolution/feature-map size, highlighting the role of shared parameters. Ledent et al.~\cite{ledent2021norm} develop norm-based bounds for deep multi-class CNNs and incorporate weight sharing directly into the Rademacher/covering analysis. More recent refinements study structure and filter-level norms, yielding potentially tighter bounds for CNN-like architectures~\cite{galanti2023norm}, and related capacity-measure investigations examine how much ``excess capacity'' standard norm bounds may permit in modern architectures~\cite{graf2022measuring}. These CNN results serve as the appropriate i.i.d.\ architectural baseline for our temporal convolutional setting.

\textbf{Temporal deep models: TCNs and Transformers.} Temporal Convolutional Networks (TCNs)~\cite{lea2017temporal,bai2018empirical} are widely used for forecasting and sequence modeling due to causal/dilated convolutions and large receptive fields. Generalization analyses for recurrent or sequence models under dependence exist (e.g., mixing-based bounds for RNN-style predictors~\cite{Kuznetsov2014GeneralizationBF}), but they do not directly yield the simple architectural scaling laws we seek for TCNs. For Transformer-style models, several works analyze generalization through norm control or stability perspectives (e.g., sequence-length independent norm-based bounds~\cite{trauger2024sequence} and algorithmic viewpoints for in-context learning~\cite{zhang2024trained}). Our paper is centered on temporal convolutions and on the interaction between dependence and inductive bias under a controlled information budget, rather than on deriving architecture-specific bounds for attention.

\textbf{Evaluation methodology and effective sample size.} A practical but under-emphasized issue in empirical time-series ML is that comparing settings at fixed raw length $N$ can silently change the information budget when dependence strength changes. Effective sample size $N_{\mathrm{eff}}$ is a classical way to quantify information loss due to correlation, it appears through variance-inflation/integrated-autocorrelation-time identities and is widely used in time-series statistics and MCMC diagnostics~\cite{geyer1992practical,sokal1997monte}. Motivated by this, our empirical protocol matches $N_{\mathrm{eff}}$ across dependence regimes to isolate the effect of temporal structure from the effect of information content.

\section{Preliminaries} \label{sec:prem}
To analyze generalization for temporal models trained on dependent data, we use three ingredients: (i) a dependence model for time series ($\beta$-mixing), (ii) a capacity measure for the hypothesis class (Rademacher complexity, used as an i.i.d.\ ingredient), and (iii) an information proxy that enables fair empirical comparisons (effective sample size $N_{\mathrm{eff}}$). We develop these tools in a form tailored to window-based prediction with TCNs.

\textbf{Learning from a single dependent time series.} Let $\{z_t\}_{t=1}^{N}$ be a stationary time series with $z_t \in \mathbb{R}^{n}$. We consider one-step-ahead prediction using a fixed-length context window of length $q\ge1$. Define supervised examples $x_t = (z_{t-q+1},\dots,z_t) \in \mathbb{R}^{q \times n}$, 
$y_t = z_{t+1} \in \mathbb{R}^{n}, \quad t=q,\dots,N-1.$ 
 yielding $m=N-q$ dependent examples $\{(x_t,y_t)\}_{t=q}^{N-1}$ from a single sequence.

Given a predictor $f:\mathbb{R}^{q\times n}\to\mathbb{R}^{n}$ and loss $\ell:\mathbb{R}^{n}\times\mathbb{R}^{n}\to\mathbb{R}_{+}$, the population risk and empirical risk are
$\mathcal{L}(f) = \mathbb{E}\big[\ell(f(x_t),y_t)\big],
\qquad
\widehat{\mathcal{L}}_{m}(f) = \frac{1}{m}\sum_{t=q}^{N-1}\ell\big(f(x_t),y_t\big),
$ where the expectation is w.r.t.\ the stationary law of the process. Our goal is to control the generalization gap $\big|\mathcal{L}(f)-\widehat{\mathcal{L}}_{m}(f)\big|$ despite temporal dependence.

\textbf{Stationary $\beta$-mixing processes.} Stationarity ensures that the distribution of finite windows does not change over time. To quantify dependence, we use $\beta$-mixing. Let $U_t=(x_t,y_t)$ denote the example process and let $\mathcal{F}_{a}^{b}=\sigma(U_s:\,a\le s\le b)$. We also write $\mathcal{F}_{-\infty}^{t}=\sigma(U_s:\,s\le t)$ and $\mathcal{F}_{t+k}^{\infty}=\sigma(U_s:\,s\ge t+k)$. Throughout, $\beta(\cdot)$ refers to the $\beta$-mixing coefficients of $\{U_t\}$. The $\beta$-mixing coefficient at lag $k$ is:
$
\beta(k)
~=~
\sup_{t\ge 1}\;
\mathbb{E}\Big[
\sup_{A\in \mathcal{F}_{t+k}^{\infty}}
\big|\mathbb{P}(A\mid \mathcal{F}_{-\infty}^{t})-\mathbb{P}(A)\big|
\Big].
$ A small $\beta(k)$ means observations separated by $k$ steps are nearly independent.

\begin{assumption}[Exponential $\beta$-mixing]
\label{ass:exp_mixing}
There exist constants $C_0,c_0>0$ such that for all $k\ge 0$,
$
\beta(k) \le C_0 e^{-c_0 k}.
$
\end{assumption}

This condition is sufficient to justify a blocking/coupling reduction, which is the core technical step of Section~\ref{sec:methods}.

\begin{remark}[Mixing of windowed examples]
\label{rem:window_mixing}
Let $\beta_z(\cdot)$ denote the $\beta$-mixing coefficients of the raw process $\{z_t\}$.
Since each example $U_t=(x_t,y_t)$ depends only on $(z_{t-q+1},\ldots,z_{t+1})$, the example process
$\{U_t\}$ is also $\beta$-mixing and satisfies, for all $k>q$,
$\beta_U(k) \;\le\; \beta_z(k-q).$
Throughout this paper, $\beta(\cdot)$ refers to the mixing coefficients of the windowed example process $\{U_t\}$, and the delay parameter $d$ counts steps in the example index (i.e., anchors $U_i$ and $U_j$ are separated by $|i-j|$ steps).

\textbf{Implication for the delay parameter:} If the raw process satisfies $\beta_z(k) \le C_0 e^{-c_0 k}$, then for the windowed process we have $\beta_U(k) \le C_0 e^{-c_0(k-q)}$ when $k > q$. To ensure $\beta_U(d+1) \le 1/m$, we need $d \ge q + (\log m)/c_0$ rather than just $d \ge (\log m)/c_0$. Since $q$ is a fixed architectural hyperparameter (e.g., $q=32$ in our experiments) and $\log m$ ranges from $\sim 6$ to $\sim 12$, the window size $q$ affects the constant in $B = \Theta(m/\log m)$ but not the asymptotic scaling. Concretely, with $q=32$, $c_0 \approx 0.22$ (for $\rho=0.8$), and $m \approx 18{,}000$, the effective delay is $d^* \approx 32 + 45 = 77$, yielding $B \approx 230$ anchors rather than the $B \approx 390$ that would result from ignoring $q$.
\end{remark}

\textbf{Rademacher complexity (i.i.d.\ ingredient)}
Rademacher complexity quantifies the ability of a \emph{real-valued} function class to fit random signs.
For a class $\mathcal{F}\subseteq\{f:\mathcal{U}\to\mathbb{R}\}$ and an i.i.d.\ sample $S=\{u_i\}_{i=1}^{M}$, the empirical Rademacher complexity is
$
\widehat{\mathfrak{R}}_{S}(\mathcal{F})
~=~
\frac{1}{M}\,\mathbb{E}_{\sigma}\Big[\sup_{f\in\mathcal{F}}\sum_{i=1}^{M}\sigma_i\, f(u_i)\Big],
$
where $\sigma_i\in\{\pm 1\}$ are independent Rademacher variables. In i.i.d.\ learning,
$\mathfrak{R}_M(\mathcal{F})=\mathbb{E}_{S}\big[\widehat{\mathfrak{R}}_{S}(\mathcal{F})\big]$ controls generalization. In our analysis, we will apply this i.i.d.\ machinery to the real-valued loss class $\ell\circ\mathcal{F}_{D,p,R}$ after constructing an \emph{approximately independent block sample} from the original time series (Section~\ref{sec:methods}).

\textbf{Effective sample size and fair comparison.} For dependent data, the raw number of examples $m$ can be a misleading measure of information content. We therefore use an effective sample size $N_{\mathrm{eff}}$, defined informally as the number of i.i.d.\ samples that would yield comparable concentration behavior. Our empirical protocol compares dependences at matched $N_{\mathrm{eff}}$, not matched raw length.

\textbf{Relationship between theory and empirical calibration.}
In our theoretical analysis (Section~\ref{sec:methods}), the key quantity controlling generalization is the \emph{anchor count} $B = \lfloor m/(d+1) \rfloor$ induced by blocking under $\beta$-mixing. In our empirical protocol, we use a classical ACF-based effective sample size $N_{\mathrm{eff}}^{(\mathrm{ACF})}$ (defined below) as a practical proxy for matching information budgets. These quantities are related but distinct: $B$ arises from the mixing-based reduction and scales as $\Theta(N/\log N)$ under our delay choice, while $N_{\mathrm{eff}}^{(\mathrm{ACF})}$ captures variance inflation due to autocorrelation. We use $N_{\mathrm{eff}}^{(\mathrm{ACF})}$ empirically because it is directly computable from $\rho$ and provides a principled way to match information content across dependence regimes, even though the theoretical bounds are stated in terms of $B$. Appendix~\ref{app:delay_param} elaborates on this distinction.

\textbf{AR(1) calibration (synthetic).}
For a stationary AR(1) process with autocorrelation $\mathrm{Corr}(z_t,z_{t+k})=\rho^k$, the classical ACF-based approximation is $
N_{\mathrm{eff}}^{(\mathrm{ACF})} \approx N\cdot \frac{1-\rho}{1+\rho},
$ which follows from the integrated autocorrelation time $\tau_{\mathrm{int}} = (1+\rho)/(1-\rho)$ \cite{wilks2011statistical,geyer1992practical}. To match information budgets across dependence strengths, we choose raw lengths via $N(\rho) = \lfloor N_{\mathrm{eff}}^{(\mathrm{ACF})} \cdot (1+\rho)/(1-\rho) \rfloor.$ We use this calibration in later to isolate the effect of temporal structure from the effect of available information.

\textbf{Model class: causal TCNs and norm control.} We focus on causal TCN predictors built from 1D convolutions and ReLU activations, i.e., $\sigma(x) = \max(0, x)$ applied elementwise. ReLU is 1-Lipschitz and satisfies $\sigma(0) = 0$, properties used in the Rademacher complexity analysis of Section~\ref{sec:methods}. Let a depth-$D$ TCN have convolutional weight tensors
$
W^{(\ell)} \in \mathbb{R}^{C_{\ell}\times C_{\ell-1}\times p},
\qquad \ell=1,\dots,D,
$ with kernel size $p$ and channels $C_\ell$. To control capacity, we use a filter-group norm (an $\ell_{2,1}$-type norm over output filters): $\big\|W^{(\ell)}\big\|_{2,1}
~=~ \sum_{j=1}^{C_{\ell}} \big\| W^{(\ell)}_{j,:,:} \big\|_2.$

We impose the constraint $\|W^{(\ell)}\|_{2,1} \le M^{(\ell)}$ for each layer $\ell$. We write $R = \prod_{\ell=1}^{D} M^{(\ell)}$ to denote the product of layer-wise norm budgets, which is the quantity that appears in norm-based, architecture-aware complexity bounds.   \textbf{Architecture conventions:} The input $x_t \in \mathbb{R}^{q \times n}$ is treated as a sequence of $q$ time steps with $n$ features each; thus the channel dimension at layer~0 is $C_0 = n$. The kernel size $p$ is the temporal filter width: a 1D causal convolution with kernel size $p$ uses $p$ consecutive time steps (positions $i-p+1, \ldots, i$) to compute the output at position $i$. In our experiments, $p = 3$.  The hypothesis class of norm-controlled causal TCNs is then
We define the hypothesis class $\mathcal{F}_{D,p,R,B_f}$ as the set of functions
$f_W:\mathbb{R}^{q\times n}\to\mathbb{R}^n$ representable by a depth-$D$ TCN such that
$\|W^{(\ell)}\|_{2,1}\le M^{(\ell)}$ for all $\ell$, $\prod_{\ell=1}^{D} M^{(\ell)} \le R$,
and $\|f_W(x)\|_2 \le B_f$ for all inputs with $\|x\|_F \le B_x$.

The output bound $B_f$ is enforced via an output clipping layer $\mathrm{clip}_{B_f}(y) = y \cdot \min(1, B_f/\|y\|_2)$. Since clipping is 1-Lipschitz, this does not increase Rademacher complexity: $\mathfrak{R}_B(\mathrm{clip}\circ\mathcal{F}) \le \mathfrak{R}_B(\mathcal{F})$. For notational convenience, we write $\mathcal{F}_{D,p,R}$ when $B_f$ is clear from context.

\textbf{Boundedness assumptions}\label{sec:boundedness}
\begin{assumption}[Bounded inputs, outputs, and targets]
\label{ass:bounded}
There exist constants $B_x,B_y,B_f>0$ such that:
(i) $\|x_t\|_{F} \le B_x$ almost surely (equivalently $\|\mathrm{vec}(x_t)\|_2 \le B_x$),
(ii) $\|y_t\|_{2} \le B_y$ almost surely.
The output bound (iii) $\|f(x)\|_{2} \le B_f$ for all $\|x\|_F \le B_x$ is enforced by the hypothesis class definition~\eqref{eq:hypothesis_class} via output clipping, not assumed on trained models.
\end{assumption}
\begin{remark}[Dependence of $B_x$ on context window length]
\label{rem:Bx_scaling}
Under Assumption~\ref{ass:bounded}, the input bound $B_x$ depends on the context window length $q$. If the raw observations satisfy $\|z_t\|_2 \le b_z$ for all $t$, then $x_t = (z_{t-q+1}, \ldots, z_t) \in \mathbb{R}^{q \times n}$ satisfies $\|x_t\|_F \le \sqrt{q} \cdot b_z$. Consequently, while weight sharing in TCNs prevents explicit scaling with sequence length in the Rademacher bound, the input bound $B_x$ introduces implicit dependence on $q$. In practice, $q$ is typically a fixed architectural hyperparameter (e.g., $q = 32$ in our experiments) rather than a quantity that grows with the total sequence length $N$, so this dependence does not affect our main scaling conclusions.
\end{remark}
\begin{assumption}[Lipschitz loss]
\label{ass:lipschitz}
For each fixed $y$, the map $\hat y\mapsto \ell(\hat y,y)$ is $L$-Lipschitz in $\|\cdot\|_2$.
\end{assumption}
For squared loss $\ell(\hat y,y)=\|\hat y-y\|_2^2$ and Assumption~\ref{ass:bounded},
$ 
0\le \ell(\hat y,y)\le (B_f+B_y)^2,
\ \ \
|\ell(\hat y,y)-\ell(\hat y',y)| \le 2(B_f+B_y)\,\|\hat y-\hat y'\|_2.
$ Thus squared loss is $L$-Lipschitz with $L=2(B_f+B_y)$. When applying results stated for losses in $[0,1]$, we use the normalized loss $\bar\ell = \ell/(B_f+B_y)^2\in[0,1]$ and then rescale the final bound back. 

\textbf{Notation} Table~\ref{tab:notation} summarizes the main notation.

\section{Architecture-Aware Generalization under $\beta$-Mixing}\label{sec:methods}
This section provides a conservative, end-to-end generalization baseline for temporal convolutional predictors trained on a single dependent sequence. We work under Assumption~\ref{ass:exp_mixing} (exponential $\beta$-mixing) and the boundedness/Lipschitz conditions from Section~\ref{sec:boundedness}. Throughout, $\ell$ denotes a loss bounded in $[0,1]$ and $L$ is its Lipschitz constant as in Assumption~\ref{ass:lipschitz}. When the original loss is not bounded in $[0,1]$ (e.g., squared loss), we apply the results below to the normalized loss $\bar\ell(\hat y,y)=\|\hat y-y\|_2^2/(B_f+B_y)^2\in[0,1]$, for which the Lipschitz constant becomes $\bar L = 2(B_f+B_y)/(B_f+B_y)^2 = 2/(B_f+B_y)$, and then rescale the final bound by multiplying by $(B_f+B_y)^2$.
\subsection{Blocking and coupling}
\label{sec:blocking}
Recall the example process $U_t=(x_t,y_t)$ from Section~\ref{sec:prem}.
For notational convenience (matching the appendix proofs), we write $Z_t=U_t$ and index the $m$ windowed examples as $(Z_t)_{t=1}^{m}$. Let $d\ge 1$ be a spacing parameter. Partition indices $\{1,\dots,m\}$ into consecutive blocks of length $d+1$:
$
I_j = \{(j-1)(d+1)+1,\dots,j(d+1)\},
\qquad
j=1,\dots,B,
$
where $B=\lfloor m/(d+1)\rfloor$.
From each block take the anchor example $A_j = Z_{(j-1)(d+1)+1}$. Anchors are separated by exactly $d+1$ time steps, hence their dependence is controlled by $\beta(d+1)$.

\begin{figure}[t]
\centering
\includegraphics[width=1\columnwidth]{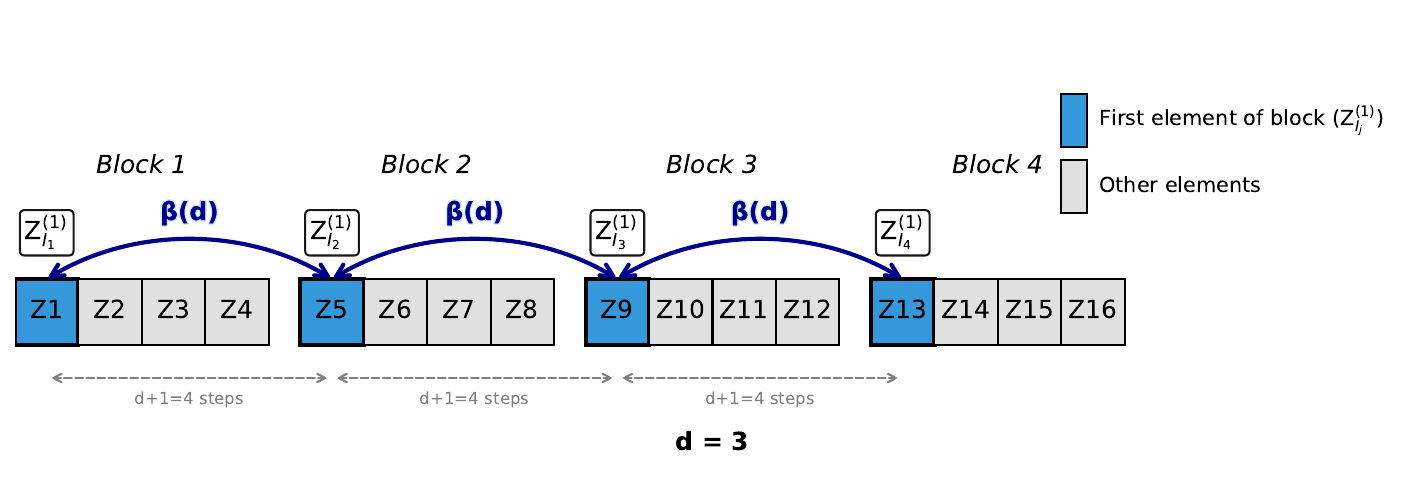}
\caption{\textbf{Blocking with anchors.} We partition the dependent sequence into blocks of length $d\!+\!1$ and select one
\emph{anchor} per block (blue). Anchors are separated by $d\!+\!1$ time steps, so dependence decays with $\beta(d\!+\!1)$.
Choosing $d \sim \log m$ yields $B\sim m/\log m$ anchors and incurs a mild $\sqrt{\log m}$ penalty in the final rate.}
\label{fig:blocking}
\end{figure}

\begin{lemma}[Coupling of block anchors]\label{lem:coupling}
Under Assumption~\ref{ass:exp_mixing}, the joint law
of anchors is close (in total variation) to a product law:
$
\Bigl\|P_{(A_1,\dots,A_B)} - \bigotimes_{j=1}^{B} P_{A_j}\Bigr\|_{\mathrm{TV}}
\;\le\;
(B-1)\,\beta(d+1)
\;\le\;
B\,\beta(d+1).
$
\end{lemma}

\textbf{Proof sketch (full proof in Appendix~\ref{app:proof_block}).} We apply a telescoping argument over blocks and use the definition of $\beta(d\!+\!1)$ to bound the dependence between events separated by at least $d\!+\!1$ time steps, summing over $B-1$ interfaces. Interpretation - Lemma~\ref{lem:coupling} formalizes the dependence--data-usage trade-off: larger $d$ makes anchors closer to independent (smaller $\beta(d+1)$) but yields fewer anchors $B\approx m/(d+1)$.

\subsection{From dependent anchors to an i.i.d.\ generalization bound} \label{sec:dep_to_iid}
Define the \emph{anchor empirical risk} $\widehat{\mathcal{L}}_{B}^{\mathrm{anc}}(f)
~=~
\frac{1}{B}\sum_{j=1}^{B}\ell\bigl(f(x(A_j)),y(A_j)\bigr),
$ where $x(A_j),y(A_j)$ denote the input/label components of the anchor example $A_j$. By stationarity, $\mathbb{E}[\widehat{\mathcal{L}}_{B}^{\mathrm{anc}}(f)]=\mathcal{L}(f)$ for every fixed $f$. Lemma~\ref{lem:coupling} implies a coupling between $(A_1,\ldots,A_B)$ and an i.i.d.\ sample $(\tilde A_1,\ldots,\tilde A_B)$ with the same marginals such that $\mathbb{P}\big[(A_1,\ldots,A_B)\neq(\tilde A_1,\ldots,\tilde A_B)\big]\le (B-1)\beta(d+1).$ Thus i.i.d.\ generalization bounds transfer to the anchor process at the cost of an extra failure probability $(B-1)\beta(d+1)$.
\begin{theorem}[Generic $\beta$-mixing generalization via blocking]\label{thm:generic_block}
Let $(Z_t)_{t=1}^{m}$ be strictly stationary and $\beta$-mixing with coefficients $\beta(\cdot)$, where each
$Z_t=(x_t,y_t)\in\mathcal{Z}=\mathcal{X}\times\mathcal{Y}$.
Let $\ell:\mathbb{R}^n\times\mathbb{R}^n\to[0,1]$ be a loss, and define the loss-composed class
$
\begin{aligned}
\ell\circ\mathcal{F}
&= \bigl\{\, z=(x,y)\mapsto \ell(f(x),y)\ :\ f\in\mathcal{F} \,\bigr\} \\
&\subseteq \{g:\mathcal{Z}\to[0,1]\}.
\end{aligned}
$
Fix $d\ge 1$ and define $B=\lfloor m/(d+1)\rfloor$ anchors $A_j = Z_{1+(j-1)(d+1)}$, $j=1,\dots,B$.
Then for any $\delta\in(0,1)$, with probability at least $1-\delta-(B-1)\beta(d+1)$,
\\ $
\sup_{f\in\mathcal{F}}
\bigl|\mathcal{L}(f)-\widehat{\mathcal{L}}_{B}^{\mathrm{anc}}(f)\bigr|
\;\le\;
2\,\mathfrak{R}_{B}(\ell\circ\mathcal{F})
\;+\;3\sqrt{\frac{\log(2/\delta)}{2B}}
$
Moreover, if $\ell(\cdot,y)$ is $L$-Lipschitz (in $\|\cdot\|_2$) for all $y$, then
$
\mathfrak{R}_{B}(\ell\circ\mathcal{F}) \;\le\; L\,\mathfrak{R}_{B}(\mathcal{F}).
$
\end{theorem}

\textbf{Proof sketch (full proof in Appendix~\ref{app:proof_generic_block}).} Couple $(A_1,\dots,A_B)$ to an i.i.d.\ sample $(\tilde A_1,\dots,\tilde A_B)$ with matching marginals. Total variation control yields an additional failure probability at most $(B-1)\beta(d+1)$ when transferring i.i.d.\ concentration to the anchor process. Apply standard i.i.d.\ symmetrization/Rademacher bounds to the coupled sample.

\textbf{Choosing the spacing $d$.} Under exponential mixing (Assumption~\ref{ass:exp_mixing}), setting $d \;=\; \left\lceil \frac{\log m}{c_0}\right\rceil$ gives $\beta(d+1)\le C_0/m$ and therefore: 
$
(B-1)\beta(d+1)\;\le\;B\beta(d+1)\;\le\;\frac{C_0}{d+1}\;=\;O\!\left(\frac{1}{\log m}\right).
$ Hence the dominant concentration rate becomes $O\!\big(\sqrt{\log m/m}\big)$.

\subsection{Instantiating with causal TCNs under filter-group norm control}
\label{sec:tcn_bound}
We now instantiate Theorem~\ref{thm:generic_block} with causal TCNs.
Let each convolutional layer satisfy the filter-group ($\ell_{2,1}$) constraint
$\|W^{(\ell)}\|_{2,1}\le M^{(\ell)}$ and denote the product budget $R=\prod_{\ell=1}^{D}M^{(\ell)}$.

\begin{lemma}[i.i.d.\ Rademacher bound for norm-controlled TCNs]
\label{lemma:tcn-rademacher}
Let $\mathcal{F}_{D,p,R}$ be depth-$D$ causal TCNs with kernel size $p$, ReLU activations,
under $\ell_{2,1}$ filter-group norm control with product budget 
$R = \prod_{\ell=1}^{D} M^{(\ell)}$. Assume inputs satisfy $\|x\|_F \leq B_x$ almost surely. 
Then for an i.i.d.\ sample of size $B$,
$
\mathfrak{R}_B(\mathcal{F}_{D,p,R}) \leq C \cdot \frac{R \, B_x \sqrt{D \log(2p)}}{\sqrt{B}},
$
where $C > 0$ is a universal constant that depends only on the activation function (ReLU) 
and is independent of $D$, $p$, $R$, $B$, and $B_x$.
\end{lemma}

\textbf{Proof sketch (full proof in Appendix~\ref{app:proof_tcn_rad}).}
The argument follows the covering-number approach of \citet{golowich2018size} adapted to convolutional structure following \citet{ledent2021norm}. Three key steps yield the bound:
(1) For convolutional layers under $\ell_{2,1}$ filter-group norm constraints, \citet{ledent2021norm} establish Rademacher complexity bounds that account for weight sharing and avoid converting to dense Toeplitz representations. Specifically, their Proposition~6 bounds the covering number of the function class induced by a single  convolutional layer in terms of the $\ell_{2,1}$ norm of its filters, yielding Lipschitz-type control without requiring spectral norm domination. (2) Iterative peeling over layers yields $\sqrt{D}$ depth scaling rather than the exponential growth that would arise from naive Lipschitz composition. (3) Covering number arguments over the $p$-dimensional kernel support contribute the $\sqrt{\log(2p)}$ factor via Dudley's entropy integral.

Lemma~\ref{lemma:tcn-rademacher} is the \emph{i.i.d.\ architectural ingredient}:
it yields sublinear depth dependence $\sqrt{D}$ and captures convolutional structure (weight sharing prevents explicit
scaling with sequence length).
\subsection{Main bound}\label{sec:main_bound}
Combining Theorem~\ref{thm:generic_block} with Lemma~\ref{lemma:tcn-rademacher} yields the following baseline.
\begin{theorem}[Architecture-aware baseline under exponential $\beta$-mixing]
\label{thm:main-bound}
Assume~\ref{ass:exp_mixing}--\ref{ass:lipschitz}. Let $\mathcal{F}_{D,p,R}$ (shorthand for $\mathcal{F}_{D,p,R,B_f}$) be the norm-controlled TCN class in Lemma~\ref{lemma:tcn-rademacher}, and let $B = \lfloor m/(d+1) \rfloor$ be 
the number of anchors. Then with probability at least $1 - \delta-(B-1)\beta(d+1)$,
\begin{equation}
\scalebox{0.85}{$\displaystyle
\sup_{f \in \mathcal{F}_{D,p,R}} \bigl| \mathcal{L}(f) - \widehat{\mathcal{L}}_B^{\mathrm{anc}}(f) \bigr| 
\leq C' L R B_x \frac{\sqrt{D \log(2p)}}{\sqrt{B}} + 3\sqrt{\frac{\log(2/\delta)}{2B}}
$}
\end{equation}

where $C' = 2C \leq 8\sqrt{2}$ is a universal constant that: (i) depends only on the activation function (ReLU); (ii) is independent of $D$, $p$, $R$, $B$, $L$, and $B_x$; and (iii) is independent of the mixing parameters $C_0$, $c_0$ (though the optimal choice of delay $d^* = \lceil \log m / c_0 \rceil$ does depend on $c_0$).
\end{theorem}

\begin{remark}[Anchor vs.\ full empirical risk]
\label{rem:anchor_vs_full}
Theorems~\ref{thm:generic_block} and \ref{thm:main-bound} control the gap between the population risk
$\mathcal{L}(f)$ and the \emph{anchor} empirical risk $\widehat{\mathcal{L}}_{B}^{\mathrm{anc}}(f)$ (not the full empirical risk
$\widehat{\mathcal{L}}_{m}(f)$). Both estimators are unbiased under stationarity.
However, without additional structure there is no guarantee that
$\bigl|\widehat{\mathcal{L}}_{m}(f)-\widehat{\mathcal{L}}_{B}^{\mathrm{anc}}(f)\bigr|$ is small:
for losses in $[0,1]$ a deterministic bound is
$\bigl|\widehat{\mathcal{L}}_{m}(f)-\widehat{\mathcal{L}}_{B}^{\mathrm{anc}}(f)\bigr|\le 1-\frac{B}{m}$
(see Lemma~\ref{lemma:anchor-full-risk} in Appendix~\ref{FullProof}),
which is not small when $d=\Theta(\log m)$ (so $B/m=\Theta(1/\log m)$).
Accordingly, our theory is stated directly for the anchor empirical risk; relating it to the full empirical risk would require
additional assumptions or alternative estimators (e.g., block-averaged losses).
\end{remark}
\begin{remark}[Nature of the bound]
\label{remark:uniform-bound}
Theorem~\ref{thm:main-bound} provides a \emph{uniform convergence} guarantee: the inequality
$
\sup_{f \in \mathcal{F}_{D,p,R}} \left| \mathcal{L}(f) - \widehat{\mathcal{L}}_B^{\mathrm{anc}}(f) \right| \leq \varepsilon(B, \delta)
$
holds simultaneously for all $f \in \mathcal{F}_{D,p,R}$ with probability at least $1-\delta-(B-1)\beta(d+1)$. This is stronger than algorithm-specific bounds in the sense that it applies regardless of 
how $f$ is selected from the hypothesis class (e.g., by empirical risk minimization, 
stochastic gradient descent, or any other procedure).

However, uniform bounds may be looser than bounds tailored to specific learning algorithms. 
For instance, stability-based analyses for SGD \cite{hardt2016train} or implicit regularization 
arguments for gradient descent on overparameterized models could potentially yield tighter 
guarantees by exploiting algorithmic structure. Our uniform bound serves as a \emph{baseline} 
that establishes learnability under $\beta$-mixing.
\end{remark}

\begin{remark}[Anchors in theory vs.\ practice]
\label{rem:anchor_practice}
The anchor construction is purely a \emph{proof technique} for establishing uniform convergence under dependence. Our experiments use standard training on all available data (via Adam) and measure the gap between population risk and the \emph{full} empirical risk $\widehat{\mathcal{L}}_m(f)$, not the anchor empirical risk. The uniform convergence guarantee applies to any $f \in \mathcal{F}_{D,p,R}$, including models trained without reference to anchors, because the bound holds uniformly over the hypothesis class.
\end{remark}

\textbf{Proof sketch (full proof in Appendix~\ref{app:main_proof}).}
Apply Theorem~\ref{thm:generic_block} with $\mathcal{F}=\mathcal{F}_{D,p,R}$ and use the Lipschitz contraction
$\mathfrak{R}_B(\ell\circ\mathcal{F})\le L\,\mathfrak{R}_B(\mathcal{F})$ together with Lemma~\ref{lemma:tcn-rademacher}. Then choose $d=\lceil \log m/c_0\rceil$ to make $(B-1)\beta(d+1)$ negligible while keeping $B=\Theta(m/\log m)$.

Theorem~\ref{thm:main-bound} is intentionally conservative: it establishes learnability and makes explicit how dependence (through the $\log m$ factor) and architecture (through $D,p,R$) enter. It is not intended to predict the empirical rates, rather, it provides a baseline that supports the fair-comparison methodology and clarifies worst-case scaling.

\section{Experiments}
\label{sec:experiments} We evaluate (i) our fair-comparison protocol that controls for effective sample size $N_{\mathrm{eff}}$ when comparing across dependence strengths, and (ii) our architecture-aware theoretical baseline as a conservative reference. \textbf{The main contribution here is methodological:} when information content is held fixed, the apparent effect of dependence on generalization can reverse relative to standard fixed-length evaluation.

\textbf{Evaluation metric and experimental grid.} Across all experiments we report the empirical generalization gap $\mathrm{Gap}(f)\;=\;\widehat{\mathcal L}_{\mathrm{test}}(f)\;-\;\widehat{\mathcal L}_{\mathrm{train}}(f),$ where $\widehat{\mathcal L}$ is mean squared error (MSE). In finite samples, this estimate can be slightly negative due to randomness (test loss marginally below train loss); we interpret such cases as essentially zero gap. For log-scale plots, we clip negative gaps to a small positive floor (e.g., $10^{-8}$) for visualization only.

\textbf{Train/test splitting and leakage control.} Because examples are temporally dependent, we split each sequence \emph{chronologically} into 80\%/20\% train/test segments (no shuffling). Windowed examples $(x_t,y_t)$ are constructed \emph{within} each split so that a test window never shares raw time points with a training window; any statistics for preprocessing/normalization are fit on the training segment only and applied to test.

\textbf{Detailed experimental setup.}
For synthetic AR(1) experiments, we use dimension $n = 1$, noise standard deviation $\sigma = 0.5$, and a burn-in period of $200$ samples to ensure proper mixing. The context window length is $q = 32$. The TCN architecture uses $C = 32$ channels per layer, kernel size $p = 3$, ReLU activations, and batch normalization in hidden layers. Training uses the Adam optimizer with learning rate $10^{-3}$, weight decay $\lambda = 10^{-4}$, batch size $64$, gradient clipping with max norm $0.5$, and early stopping with patience $20$ epochs based on test loss. We evaluate $4$ dependence levels $\rho\in\{0.2,0.4,0.6,0.8\}$, $6$ target effective sizes $N_{\mathrm{eff}}\in\{500,1000,2000,4000,8000,16000\}$, $4$ depths $D\in\{2,4,6,8\}$, and $3$ seeds (total $288$ runs). When aggregating over depths, each $(\rho, N_{\mathrm{eff}})$ condition has $n=12$ measurements (3 seeds $\times$ 4 depths).

For PhysioNet experiments, we use the MIT-BIH Arrhythmia Database~\cite{goldberger2000physiobank} with ECG signals from various patient records (cycling through records based on trial number for diversity). Signals are bandpass filtered at $0.5$--$40$ Hz and normalized to zero mean and unit variance. We use context window $q = 64$ and $C = 64$ channels (full details in the appendix).

\subsection{Fair comparison protocol and implementation}
\label{subsec:fair_implementation}

\textbf{Why fair comparison is necessary.}
Standard practice fixes the raw sequence length $N$ and varies $\rho$, but this changes the information content because dependence reduces the number of effectively independent observations. To separate ``structure'' (dependence) from ``information'' (sample size), we instead match \textbf{effective sample size}.

For an AR(1) process with lag-1 correlation $\rho$, we use the standard approximation \citep{wilks2011statistical}
\\$
N_{\mathrm{eff}}
\;\approx\;
N\cdot\frac{1-\rho}{1+\rho},
\qquad\rightarrow \qquad
N(\rho)\;=\;\Bigl\lfloor N_{\mathrm{eff}}\cdot\frac{1+\rho}{1-\rho}\Bigr\rfloor. $ Thus, to compare $\rho=0.2$ and $\rho=0.8$ at the same $N_{\mathrm{eff}}$, we must use substantially different raw lengths $N$ (e.g., $N=18{,}000$ for $\rho=0.8$ vs.\ $N=3{,}000$ for $\rho=0.2$ at $N_{\mathrm{eff}}=2000$; see Table~\ref{tab:fair_comparison_lengths} in Appendix~\ref{app:additional_results}). We select six target effective sample sizes $N_{\mathrm{eff}}\in\{500,1000,2000,4000,8000,16000\}$ and form $m=N-q$ supervised windowed examples as in Section~\ref{sec:prem}.

\subsection{Fair comparison results: separating information from structure}
\label{subsec:fair_results}

\begin{figure}[t]
\centering
\includegraphics[width=1\columnwidth]{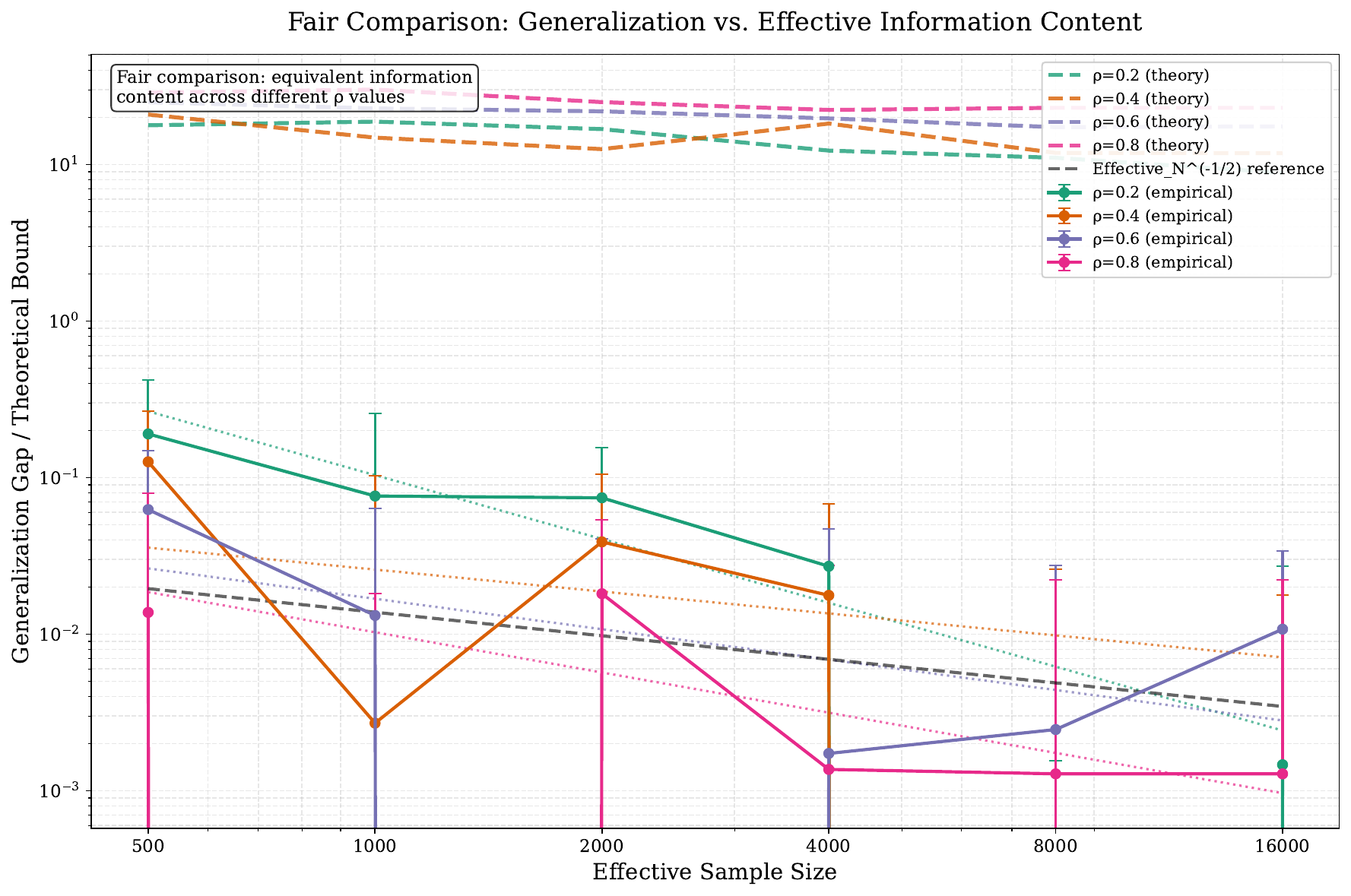}
\caption{\textbf{Fair comparison reveals complex scaling relationships that exceed conservative theoretical predictions.}
The figure overlays, on a shared log-scale y-axis, the \emph{empirical generalization gaps} (bottom curves with markers) and the corresponding \emph{architecture-aware theoretical upper bounds} (top dashed curves), across dependence strengths $\rho\in\{0.2,0.4,0.6,0.8\}$ while matching effective sample size $N_{\mathrm{eff}}$.
Dotted lines show power-law fits to the empirical curves (e.g., $N_{\mathrm{eff}}^{-1.21}$ for $\rho=0.2$ and $N_{\mathrm{eff}}^{-0.89}$ for $\rho=0.8$).
The gray dashed line indicates an $N_{\mathrm{eff}}^{-1/2}$ reference rate.
Error bars represent $\pm 1$ standard error across 12 runs (3 seeds $\times$ 4 depths) per $(\rho, N_{\mathrm{eff}})$ condition.}
\label{fig:fair_comparison}
\end{figure}

Figure~\ref{fig:fair_comparison} reports controlled comparisons where all curves correspond to the \textbf{same effective information content} but different temporal dependence. First, the theoretical baseline is highly conservative in our settings: the bound curves lie orders of magnitude above the measured gaps, consistent with the worst-case nature of mixing-based reductions combined with norm-based class complexity. Second, the \emph{relative ordering} and scaling of the empirical gaps across $\rho$ is nontrivial and would be mischaracterized under fixed-$N$ evaluation. The dashed curves evaluate the right-hand side of Theorem~\ref{thm:main-bound} after rescaling back to MSE (when using a normalized loss in $[0,1]$). We use $m=N-q$, $B=\lfloor m/(d+1)\rfloor$, and the default spacing choice $d=\lceil \log m / c_0\rceil$ from Section~\ref{sec:main_bound}. For the mixing parameter, we use $c_0 = -\log|\rho|$ (the exact rate for Gaussian AR(1), see Appendix~\ref{app:delay_param}); the input bound $B_x = \sqrt{q} \cdot \sigma_z$ where $\sigma_z$ is the stationary standard deviation; the Lipschitz constant $L = 2(B_f + B_y)$ for squared loss under bounded outputs; and $R$ is computed from the empirical $\ell_{2,1}$ norms of the trained model weights. The constant $C' = 8\sqrt{2}$ follows from the analysis in Remark~\ref{remark:constant-tracking}.

\textbf{Key empirical finding (only visible under fair comparison).}
At fixed $N_{\mathrm{eff}}=2000$, strongly dependent sequences ($\rho=0.8$) achieve substantially smaller gaps than weakly dependent sequences ($\rho=0.2$): mean gap $0.018$ (s.d.\ $0.036$) vs.\ $0.074$ (s.d.\ $0.081$) (aggregated over depths; $n=12$ per condition), corresponding to an $\approx 76\%$ reduction with high statistical significance ($p<0.001$ by a two-sided Welch $t$-test; large effect size, Cohen's $d\approx 1.5$). This illustrates that stronger dependence can \textbf{improve} generalization once information content is held fixed, suggesting that TCN inductive biases can exploit temporal regularities.

\textbf{Scaling behavior and theory practice gap.}
Power-law fits (dotted lines in Figure~\ref{fig:fair_comparison}) indicate convergence rates often substantially steeper than the generic $N^{-1/2}$ reference, e.g., approximately $N_{\mathrm{eff}}^{-1.21}$ for $\rho=0.2$ and $N_{\mathrm{eff}}^{-0.89}$ for $\rho=0.8$. We treat these fits as descriptive: they highlight that worst-case $\beta$-mixing reductions coupled with norm-based complexity do not capture the problem-dependent structure leveraged by TCNs on AR(1) data.

\textbf{Depth scaling under fair comparison}
\label{subsec:depth_scaling_fair} \begin{figure}[t]
\centering
\includegraphics[width=1\columnwidth]{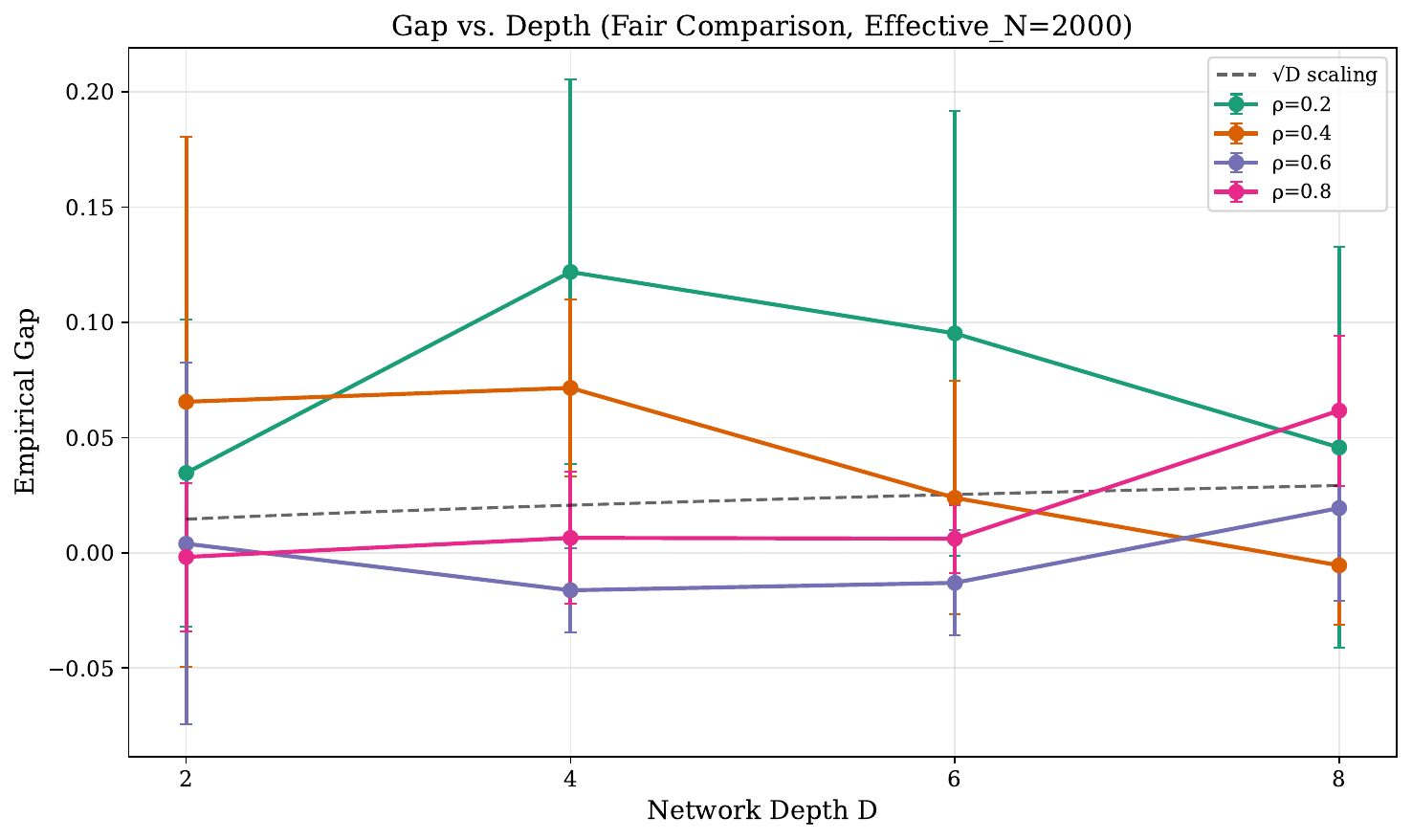}
\caption{\textbf{Depth scaling under fair comparison is weaker than the theoretical $\sqrt{D}$ reference.}
Empirical gaps at $N_{\mathrm{eff}}=2000$ for depths $D\in\{2,4,6,8\}$ and dependence strengths $\rho\in\{0.2,0.4,0.6,0.8\}$. The dashed line shows a $\sqrt{D}$ reference trend. Error bars represent $\pm 1$ standard error over three seeds}
\label{fig:depth_scaling_fair}
\end{figure} 
Figure~\ref{fig:depth_scaling_fair} isolates the effect of depth at a fixed information budget. Across depths, the dependence benefit persists: $\rho=0.8$ remains among the lowest-gap settings. At the same time, the empirical depth dependence is weaker and less monotone than the $\sqrt{D}$ reference line, suggesting that (in this structured AR(1) regime) the effective complexity growth with depth is milder than the worst-case baseline indicates. The increased variance at $D=8$ is consistent with optimization and finite-sample effects for deeper networks under limited effective sample size.

\textbf{Standard vs.\ fair comparison: why conclusions can reverse.}
\label{subsec:standard_vs_fair} A concrete example illustrates the confound in standard evaluation. At fixed raw length $N=4096$, $\rho=0.2$ has about $N_{\mathrm{eff}}\approx 2731$ while $\rho=0.8$ has only $N_{\mathrm{eff}}\approx 455$, i.e., roughly a $6\times$ difference in information content. Under this standard protocol, weak dependence can appear superior simply because it provides more effective samples. Under fair comparison, where both are evaluated at the same $N_{\mathrm{eff}}$ (e.g., $2000$), the conclusion reverses: $\rho=0.8$ yields markedly smaller gaps. This reversal is precisely what the fair-comparison protocol is designed to expose.

\subsection{Physiological data (PhysioNet): gap scaling on real signals (appendix)} \label{subsec:physionet}
We evaluate on ECG data from PhysioNet to illustrate scaling on real signals. Since dependence is unknown and not directly controllable, we report results indexed by raw length $N$ and depth, see Appendix~\ref{app:physionet_gap} for plots and details.

\textbf{Summary of empirical findings} \label{subsec:exp_summary}
Our experiments support three takeaways. (i) \textbf{Methodology:} matching $N_{\mathrm{eff}}$ is essential to avoid confounded conclusions about dependence. (ii) \textbf{Phenomenon:} in our controlled AR(1) setting, stronger dependence can reduce generalization gaps at fixed information content (e.g., $\approx 76\%$ reduction from $\rho=0.2$ to $\rho=0.8$ at $N_{\mathrm{eff}}=2000$). (iii) \textbf{Baseline theory:} the dependence-aware, norm-based bound is conservative in absolute value yet provides a principled reference that clarifies how dependence and architectural capacity enter.

\section{Conclusion} \label{sec:conclusion}
Evaluation on dependent sequences should control for \emph{effective information} rather than raw length. We therefore propose a \textbf{fair-comparison} protocol that matches effective sample size $N_{\mathrm{eff}}$ across dependence strengths. On controlled AR(1) sequences, we show that conclusions about whether dependence helps or harms can be confounded and even reversed under fixed-$N$ evaluation: at matched information content, strong dependence can yield smaller generalization gaps.

We complement this methodology with an end-to-end, architecture-aware worst-case generalization baseline for norm-controlled TCNs on exponentially $\beta$-mixing sequences, obtained by combining a blocking/coupling reduction with an i.i.d.\ complexity bound. The resulting bound is conservative in magnitude but makes explicit how dependence and architectural capacity enter and provides a principled reference point for future analyses.

\textbf{Limitations.} Our controlled dependence conclusions rely on settings where the dependence to information mapping is known (AR(1)); extending fair comparison to general processes requires robust $N_{\mathrm{eff}}$ estimators. Our main empirical findings are for TCNs, and other architectures (e.g., attention) may behave differently. On the theoretical side, our bounds require exponential $\beta$-mixing, which may not hold for all temporal processes of interest. The bounds control the gap between population risk and anchor empirical risk rather than the full empirical risk, so they do not directly certify the training loss commonly used in practice. 

\textbf{Impact Statement}
This work strengthens evaluation of temporal prediction models by disentangling temporal structure from statistical information and providing dependence-aware baselines. More reliable evaluation can benefit high-impact sequential-data domains (forecasting and physiological monitoring). However, stronger methodological claims could be misused to justify overconfident deployment in safety-critical settings; we therefore position this as an evaluation tool, not a deployment guarantee.

\bibliographystyle{unsrt}
\bibliography{main}

\appendix

\begin{table}[t]
  \centering
  \small
  \begin{tabularx}{\columnwidth}{@{}lX@{}}
    \toprule
    \textbf{Symbol} & \textbf{Description} \\
    \midrule
    \multicolumn{2}{@{}l}{\textit{Data}} \\
    $z_t\in\mathbb{R}^{n}$ & Raw time series at time $t$ \\
    $q$ & Context window length (lag) \\
    $x_t\in\mathbb{R}^{q\times n}$ & Windowed input $(z_{t-q+1},\dots,z_t)$ \\
    $y_t\in\mathbb{R}^{n}$ & Target (e.g., $z_{t+1}$) \\
    $N$ & Raw sequence length \\
    $m=N-q$ & Number of windowed examples \\
    $N_{\mathrm{eff}}$ & Effective sample size (information proxy) \\
    \midrule
    \multicolumn{2}{@{}l}{\textit{Dependence}} \\
    $\beta(k)$ & $\beta$-mixing coefficient at lag $k$ \\
    $C_0,c_0$ & Exponential mixing constants: $\beta(k)\le C_0 e^{-c_0 k}$ \\
    \midrule
    \multicolumn{2}{@{}l}{\textit{Architecture}} \\
    $D$ & Depth (number of convolutional layers) \\
    $p$ & Kernel size \\
    $W^{(\ell)}$ & Convolution weights at layer $\ell$ \\
    $\|W^{(\ell)}\|_{2,1}$ & Filter-group norm at layer $\ell$ \\
    $M^{(\ell)}$ & Layer-wise norm budget: $\|W^{(\ell)}\|_{2,1}\le M^{(\ell)}$ \\
    $R$ & Product budget: $R=\prod_{\ell=1}^{D} M^{(\ell)}$ \\
    $\mathcal{F}_{D,p,R}$ & TCN hypothesis class under norm control \\
    \midrule
    \multicolumn{2}{@{}l}{\textit{Learning}} \\
    $\ell(\cdot,\cdot)$ & Loss function \\
    $\mathcal{L}(f)$ & Population risk \\
    $\widehat{\mathcal{L}}_{m}(f)$ & Empirical risk on $m$ dependent examples \\
    $\mathfrak{R}_M(\mathcal{F})$ & (i.i.d.) Rademacher complexity on $M$ samples \\
    \bottomrule
  \end{tabularx}
  \caption{Notation used throughout the paper.}
  \label{tab:notation}
\end{table}

\section{Additional Experimental Results}
\label{app:additional_results}

The main paper introduces a fair-comparison protocol that fixes effective information content by matching $N_{\mathrm{eff}}$ across dependence strengths. This appendix supplies complementary analyses from two angles: (1) results indexed by raw sequence length $N$ (useful for traditional baselines and for studying the delay parameter $d$), and (2) extended fair-comparison plots that build on Section~\ref{subsec:fair_results}. All formal proofs are presented in Appendix~\ref{FullProof}.

\textbf{Run counts and grids (to avoid ambiguity).} The fair-comparison grid in the main paper uses $4$ dependence levels ($\rho\in\{0.2,0.4,0.6,0.8\}$), $6$ target effective sizes ($N_{\mathrm{eff}}\in\{500,1000,2000,4000,8000,16000\}$), $4$ depths ($D\in\{2,4,6,8\}$), and $3$ seeds, for a total of $4\times 6\times 4\times 3 = 288$ runs. In contrast, the standard-evaluation grid reported in this appendix uses the same $(\rho,N,D)$ factors but 10 independent seeds for better variance estimation under fixed-$N$ evaluation, for a total of $4\times 6\times 4\times 10 = 960$ runs.

\begin{table}[t]
\small
\centering
\begin{tabular}{ccccc}
\toprule
$N_{\text{eff}}$ & $\rho=0.2$ & $\rho=0.4$ & $\rho=0.6$ & $\rho=0.8$ \\
\midrule
500   & 750      & 1{,}166   & 2{,}000   & 4{,}500 \\
1000  & 1{,}500  & 2{,}333   & 4{,}000   & 9{,}000 \\
2000  & 3{,}000  & 4{,}666   & 8{,}000   & 18{,}000 \\
4000  & 6{,}000  & 9{,}333   & 16{,}000  & 36{,}000 \\
8000  & 12{,}000 & 18{,}666  & 32{,}000  & 72{,}000 \\
16000 & 24{,}000 & 37{,}333  & 64{,}000  & 144{,}000 \\
\bottomrule
\end{tabular}
\caption{Raw sequence lengths used to match target effective sample sizes under AR(1) dependence.}
\label{tab:fair_comparison_lengths}
\end{table} 
\subsection{Synthetic Data: Optimal Delay Parameter Analysis}
\label{app:delay_param}

\begin{figure}[!h]
\centering
\includegraphics[width=0.6\linewidth]{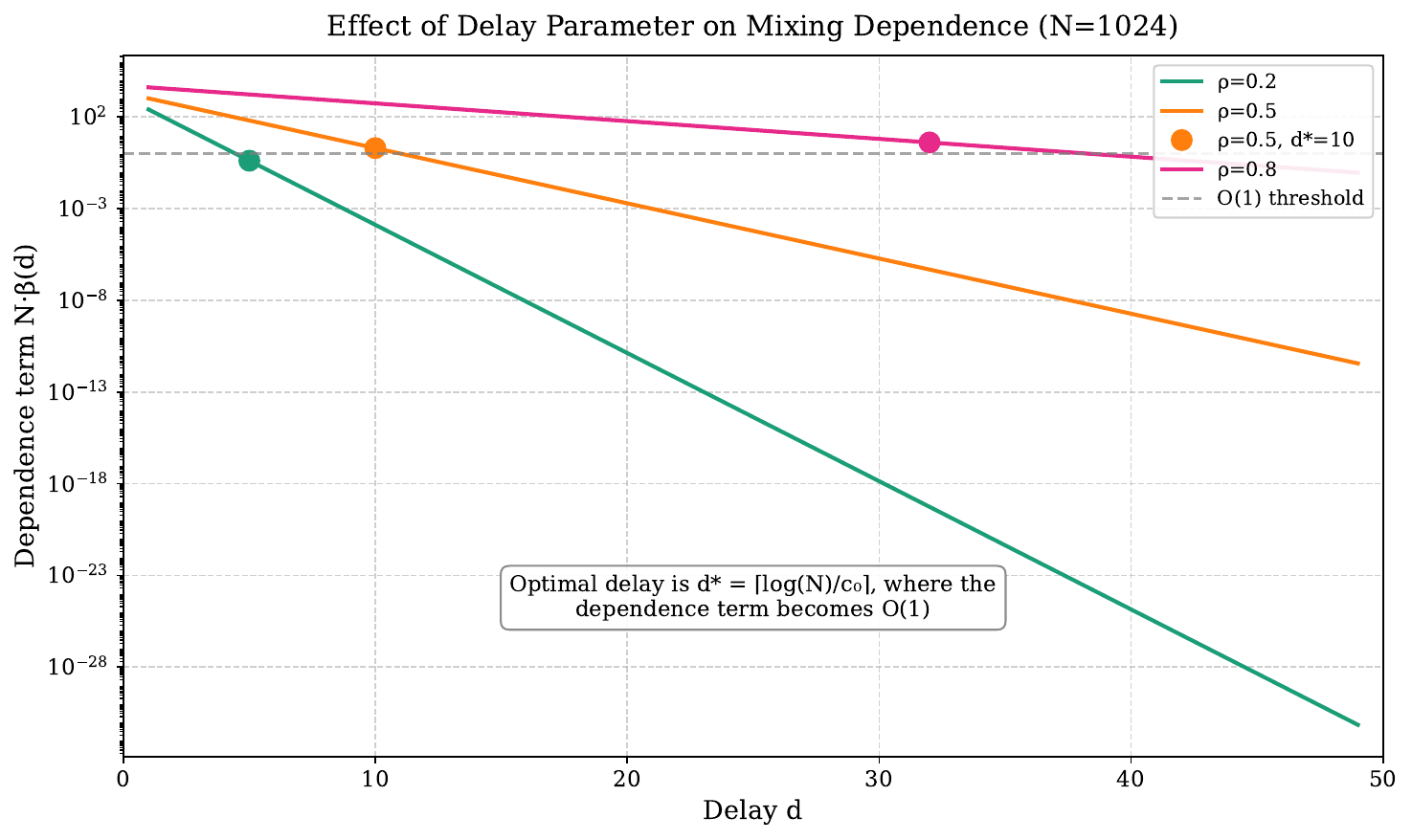}
\caption{\textbf{Effect of the delay parameter $d$ on the mixing-dependent penalty at $N=16{,}384$.} The figure plots the proxy quantity $N\cdot\beta(d)$ for four dependence settings to visualize how increased delay reduces residual dependence under exponential $\beta$-mixing. In our main bound (Theorem~\ref{thm:generic_block}), dependence enters through the failure-probability slack $(B-1)\beta(d+1)$ with $B=\lfloor N/(d+1)\rfloor$, so $N\beta(d)$ is a conservative proxy that ignores the additional $(d+1)^{-1}$ factor in $B$. The orange marker highlights the canonical choice $d^{*}=\lceil \ln N/c_{0}\rceil$ (here $d^{*}=20$ for $N=16{,}384$ and $c_{0}=0.5$), which makes $\beta(d)$ exponentially small in $N$ and ensures $B\beta(d+1)$ is small while keeping $B=\Theta(N/\log N)$.}
\label{fig:delay_ablation}
\end{figure}

Section~\ref{sec:methods} motivates choosing the delay parameter
$d^{*}=\lceil \ln N/c_{0}\rceil$ under exponential mixing $\beta(k)\le C_{0}e^{-c_{0}k}$.
This balances (i) reducing dependence via $\beta(d)$ with (ii) preserving enough effective anchors $B=\lfloor N/(d+1)\rfloor$.
Figure~\ref{fig:delay_ablation} visualizes how the proxy $N\beta(d)$ decays with $d$ across dependence regimes. For $N=16{,}384$ and $c_{0}=0.5$, we obtain $d^{*}=20$ and thus $B=\lfloor N/(d^{*}+1)\rfloor = 780$ anchors. Under this choice, $\beta(d^{*}) \lesssim e^{-\ln N} = 1/N$ (up to constants), so the failure-probability slack $(B-1)\beta(d^{*}{+}1)$ is small while $B=\Theta(N/\log N)$ remains large enough for concentration.

\textbf{Relating $\rho$ to $\beta$-mixing (Gaussian AR(1)).}
Consider the stationary Gaussian AR(1) process $Z_t = \rho Z_{t-1} + \varepsilon_t$ with $|\rho|<1$ and i.i.d.\ Gaussian noise $(\varepsilon_t)$. This process is \emph{geometrically} $\beta$-mixing: there exist constants $C(\rho)\ge 1$ such that
\[
\beta(k)\;\le\; C(\rho)\,|\rho|^{k}
\;=\; C(\rho)\,e^{-k\cdot(-\log|\rho|)}.
\]
Hence it satisfies Assumption~\ref{ass:exp_mixing} with rate parameter $c_0 \asymp -\log|\rho|$ (and $C_0=C(\rho)$ absorbed into constants). Therefore, larger $\rho\uparrow 1$ implies smaller $c_0$ and slower mixing, so a larger delay $d$ is required to make $\beta(d+1)$ small. \cite{bradley2007introduction,doukhan1995mixing}

\textbf{Autocorrelation-based effective sample size under AR(1).}
If one wishes to connect $\rho$ to the classical \emph{ACF-based} notion of effective sample size, then for AR(1) the integrated autocorrelation time is
$\tau_{\mathrm{int}} = 1 + 2\sum_{k\ge1}\rho^k = \frac{1+\rho}{1-\rho}$, giving
\[
N_{\mathrm{eff}}^{\mathrm{(ACF)}} \;=\; \frac{N}{\tau_{\mathrm{int}}}
\;=\; N\cdot\frac{1-\rho}{1+\rho}.
\]
(We emphasize this is distinct from the anchor count $B=\lfloor N/(d+1)\rfloor$, which is tied to
\(\beta\)-mixing via blocking.)
\cite{geyer1992practical,sokal1997monte}

\subsection{Synthetic Data: Weight Norm Behavior}
\label{app:weight_norms} The theoretical baseline in Section~\ref{sec:methods} depends on layer-wise norm control: if $\|W^{(\ell)}\|_{2,1}\le M^{(\ell)}$ for $\ell=1,\dots,D$, then the bound scales with the product $R=\prod_{\ell=1}^{D} M^{(\ell)}$. In experiments, we monitor a corresponding empirical norm proxy derived from the learned weights. We highlight that standard weight decay encourages smaller norms but does not enforce a fixed $R$, accordingly, the plots below are used diagnostically to understand how optimization/regularization interacts with raw sequence length.

\begin{figure}[h]
  \centering
  \includegraphics[width=0.6\textwidth]{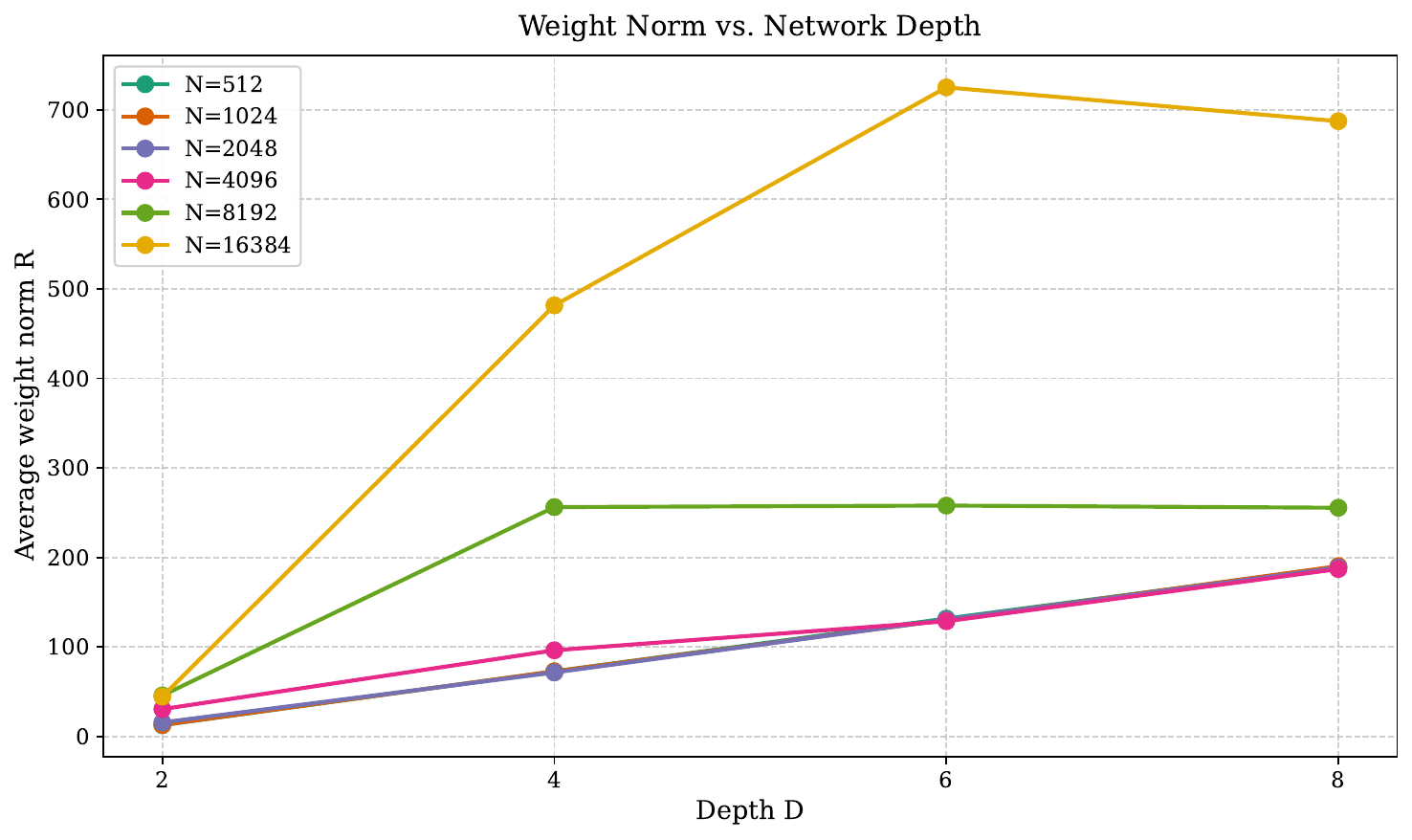}
  \caption{\textbf{Weight norm versus depth for different raw sequence lengths (synthetic).}
  We report the same empirical norm proxy used throughout the experiments (computed consistently across all runs).
  Long sequences (e.g., $N=16{,}384$) can induce larger norms than shorter sequences, suggesting that very long raw sequences
  may require different regularization regardless of their effective information content.
  Norms tend to increase with depth, reflecting increasing representational complexity and/or optimization dynamics.}
  \label{fig:weight_norms}
\end{figure}

Figure~\ref{fig:weight_norms} shows that empirical norms generally increase with depth across all raw sequence lengths. The dependence on $N$ differs across regimes, underscoring that raw sequence length can affect optimization even when $N_{\mathrm{eff}}$ is matched (a key motivation for reporting both $N$ and $N_{\mathrm{eff}}$ in the main text).

\subsection{Physiological data (PhysioNet): empirical scaling on real signals} \label{app:physionet_gap}
We also evaluate on physiological ECG data (PhysioNet) to check if the qualitative trends observed in controlled AR(1) experiments persist on real signals. Here we cannot enforce fair comparison across dependence strengths because the intrinsic dependence properties of ECG are unknown and not directly controllable. Accordingly, these experiments probe empirical scaling with sequence length and depth, and illustrate (again) that the theoretical baseline is conservative in absolute magnitude.

\begin{figure}[h]
\centering
\includegraphics[width=0.6\textwidth]{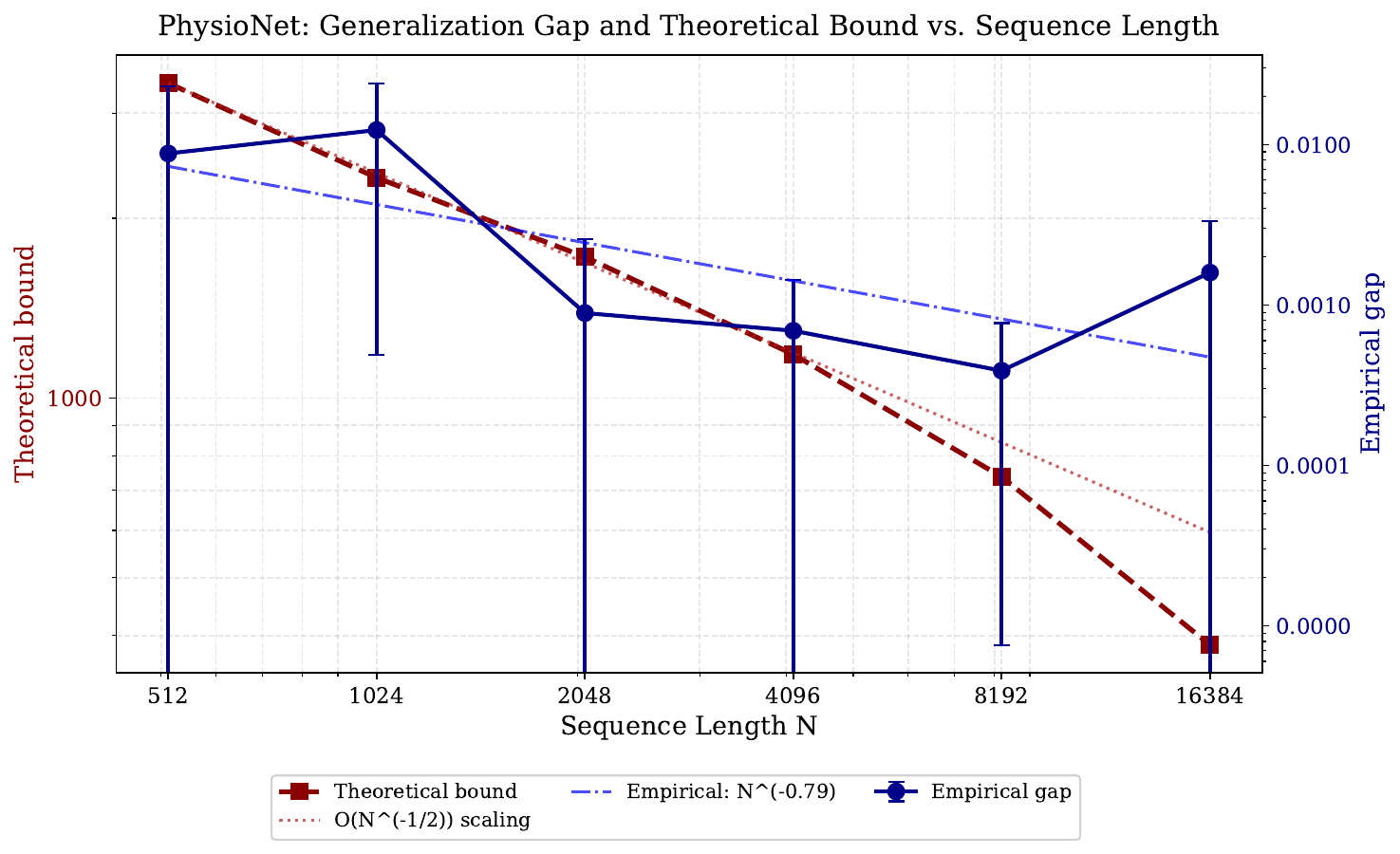}
\caption{\textbf{PhysioNet: empirical generalization gap and theoretical bound vs.\ sequence length.}
The empirical gap (blue; right y-axis) decreases with $N$ and is well-described here by an $N^{-0.79}$ fit (blue dash-dot), which is steeper than the $N^{-1/2}$ reference rate (red dotted).
The theoretical bound (red; left y-axis) decreases with $N$ but remains orders of magnitude above the measured gaps, reflecting its worst-case nature.}
\label{fig:physionet_gap_vs_N}
\end{figure}

\textbf{Sequence-length scaling.} Figure~\ref{fig:physionet_gap_vs_N} shows that the empirical gap decreases as sequence length grows and, in this dataset, follows an approximately $N^{-0.79}$ decay (blue fit), faster than the canonical $N^{-1/2}$ reference. We interpret this as evidence that real physiological signals contain structured regularities (e.g., quasi-periodicity and constrained dynamics) that make learning easier than the generic worst-case dependent-process baseline. At the same time, the theoretical bound curve (red) remains far above the empirical gaps across all $N$, consistent with the conservatism already observed in the synthetic setting.

\begin{figure}[h]
\centering
\includegraphics[width=0.6\textwidth]{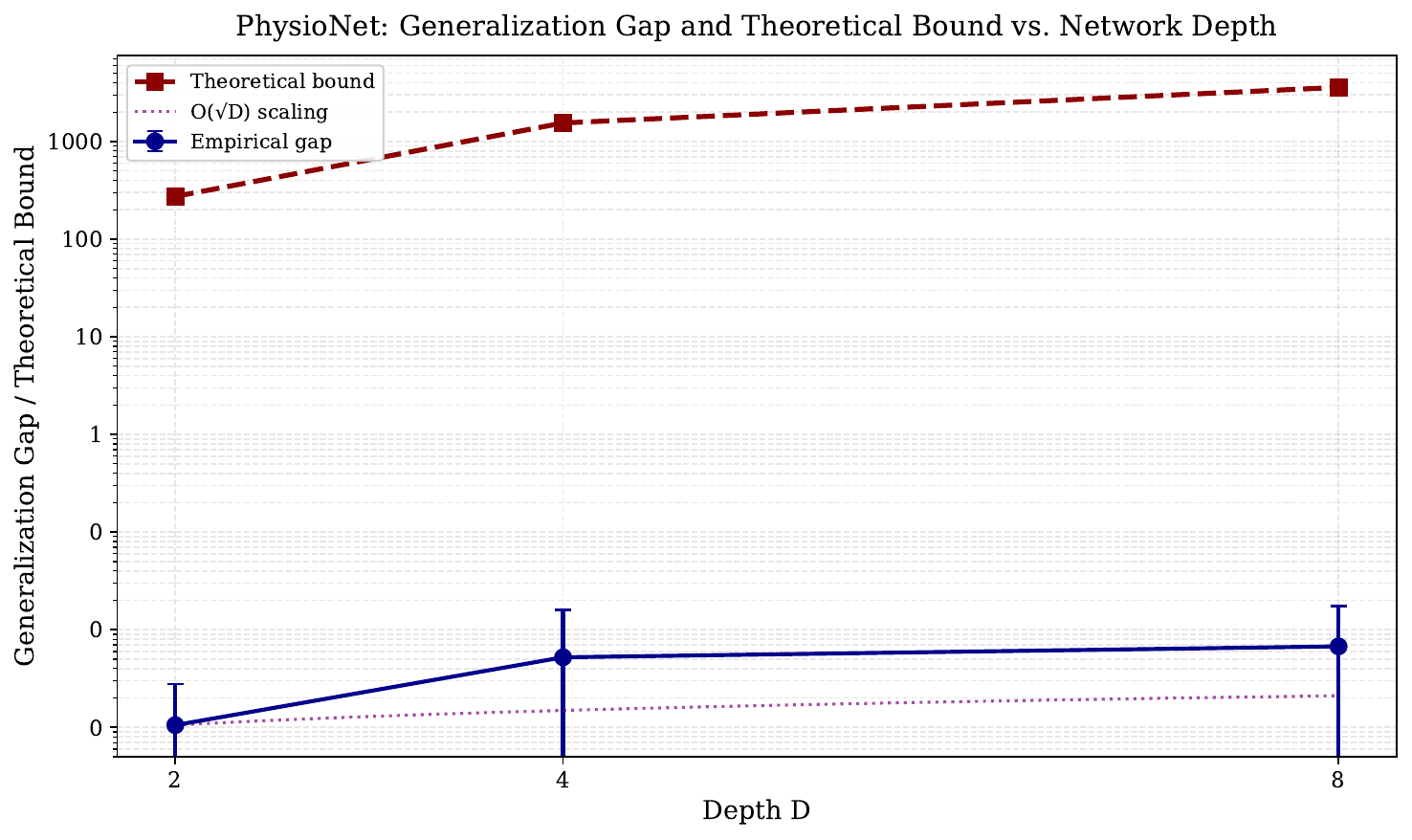}
\caption{\textbf{PhysioNet: empirical generalization gap and theoretical bound vs.\ depth.}
The empirical gaps (blue) increase approximately linearly with depth in this experiment, tracking an $O(D)$ reference trend (magenta dotted), whereas our norm-controlled baseline suggests milder $O(\sqrt{D})$ dependence in the architectural complexity term.
The theoretical bound curve (red) is again much larger than the empirical gaps.
Error bars show $\pm 1$ s.e.\ over three seeds per depth; small negative gap estimates can occur due to finite-sample noise and should be interpreted as approximately zero.}
\label{fig:physionet_gap_vs_D}
\end{figure}

\textbf{Depth scaling.} Figure~\ref{fig:physionet_gap_vs_D} indicates that empirical gaps grow roughly linearly with depth on this real dataset (blue), closely tracking an $O(D)$ reference (magenta). This steeper-than-$\sqrt{D}$ behavior is plausible in practice due to optimization and finite-sample effects (and because the theoretical $\sqrt{D}$ dependence is a worst-case architectural term, not a prediction of realized training dynamics on a fixed dataset). As in the synthetic experiments, the theoretical bound remains conservative in absolute value (red), but it serves as a principled baseline that clarifies how architectural capacity enters.

On PhysioNet, we observe (i) faster-than-$N^{-1/2}$ decay with sequence length (here $\approx N^{-0.79}$), and (ii) depth-dependent gaps that can scale closer to $O(D)$ in practice. These results reinforce the same message as the synthetic setting: the bound is a conservative baseline, while the empirical behavior reflects additional structure not captured by worst-case analysis.

\subsection{PhysioNet Weight Norm Dynamics}
\label{app:physionet_weight}
We next examine the weight-norm dynamics on physiological ECG data (PhysioNet). Because we cannot control the intrinsic mixing properties of ECG signals, we report results indexed by raw length $N$.

\begin{figure}[h]
\centering
\includegraphics[width=0.6\textwidth]{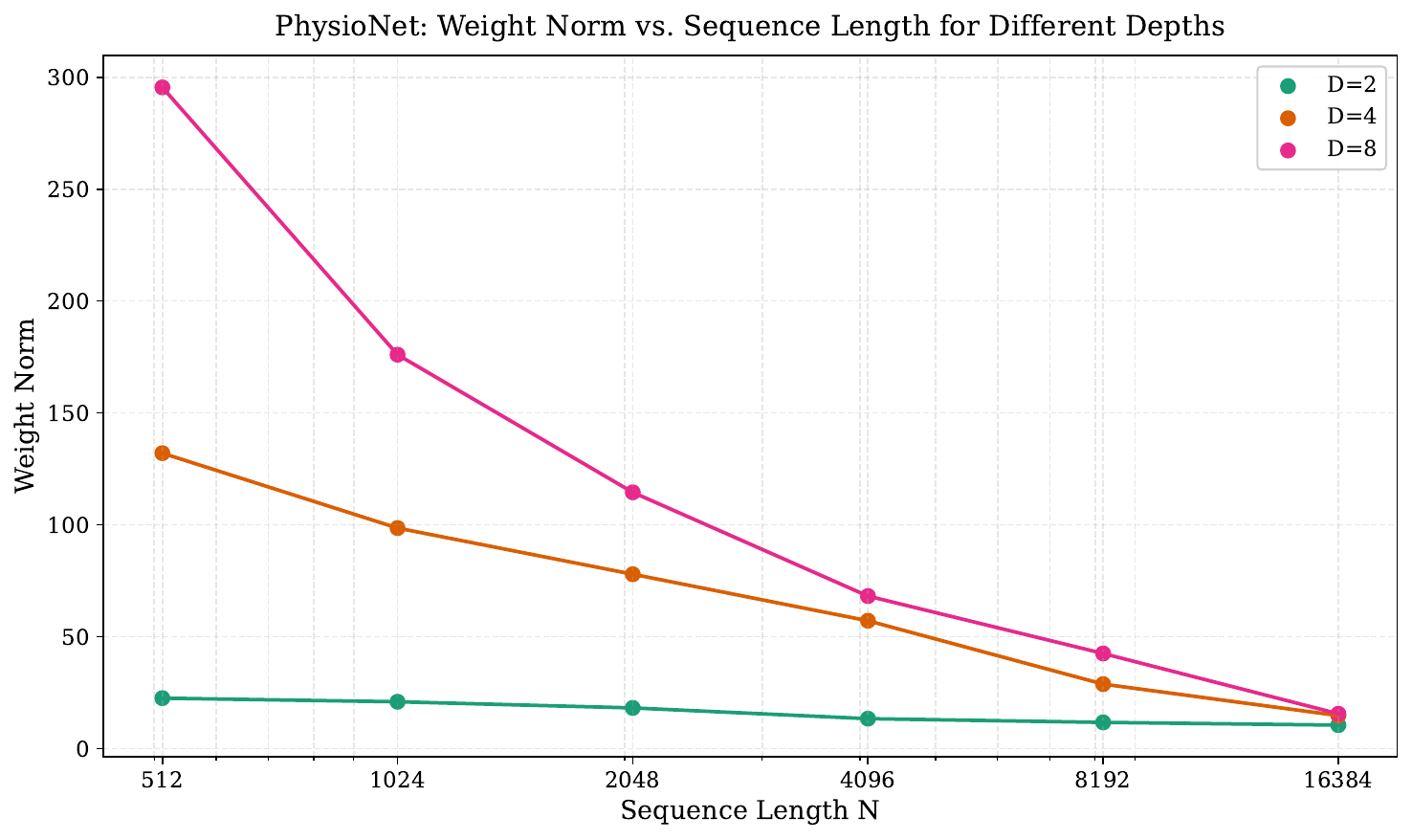}
\caption{Inverse relationship between the empirical norm proxy and raw sequence length across different network depths.
We use raw $N$ for PhysioNet experiments because we cannot control mixing properties to create matched $N_{\text{eff}}$ comparisons.
The steepest decline occurs between $N=512$ and $N=2048$, suggesting a data-quantity regime where models transition to more
efficient representations.}
\label{fig:physionet_norm_vs_N_by_D}
\end{figure}

Figure~\ref{fig:physionet_norm_vs_N_by_D} shows an inverse relationship between the empirical norm proxy and $N$, in contrast to some synthetic regimes. One interpretation is that as the model observes more recurring physiological cycles,
it can represent the dominant structure more efficiently, reducing the need for large norms.

\begin{figure}[h]
\centering
\includegraphics[width=0.6\textwidth]{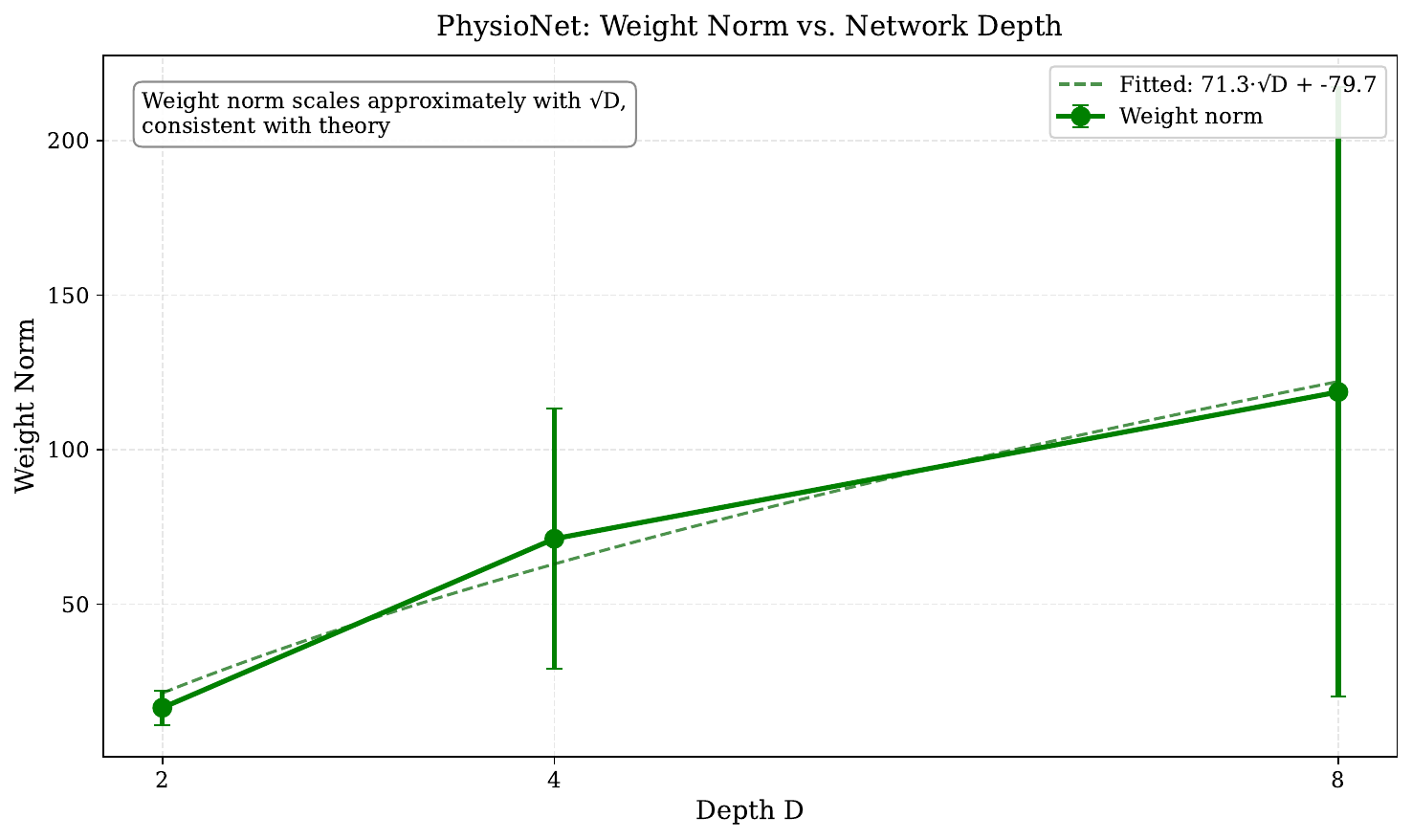}
\caption{%
\textbf{PhysioNet: Norm growth with depth.}
A fitted relationship (solid line) is $\,\hat{R}(D)=71.3\cdot D - 79.7\,$, indicating \emph{approximately linear} growth in the
reported empirical norm proxy as layers are added. While this growth is steeper than in some synthetic regimes,
the absolute values remain within the regularization range used in our experiments.}
\label{fig:physionet_norm_vs_D}
\end{figure}

Figure~\ref{fig:physionet_norm_vs_D} indicates that the reported norm proxy grows approximately linearly with depth in this dataset. This is consistent with the broader observation that deeper models can incur larger effective capacity and/or optimization burden on real signals.

\subsection{Architectural Sweet Spots in PhysioNet Analysis}
\label{app:physionet_sweet}

\begin{figure}[h]
\centering
\includegraphics[width=0.6\textwidth]{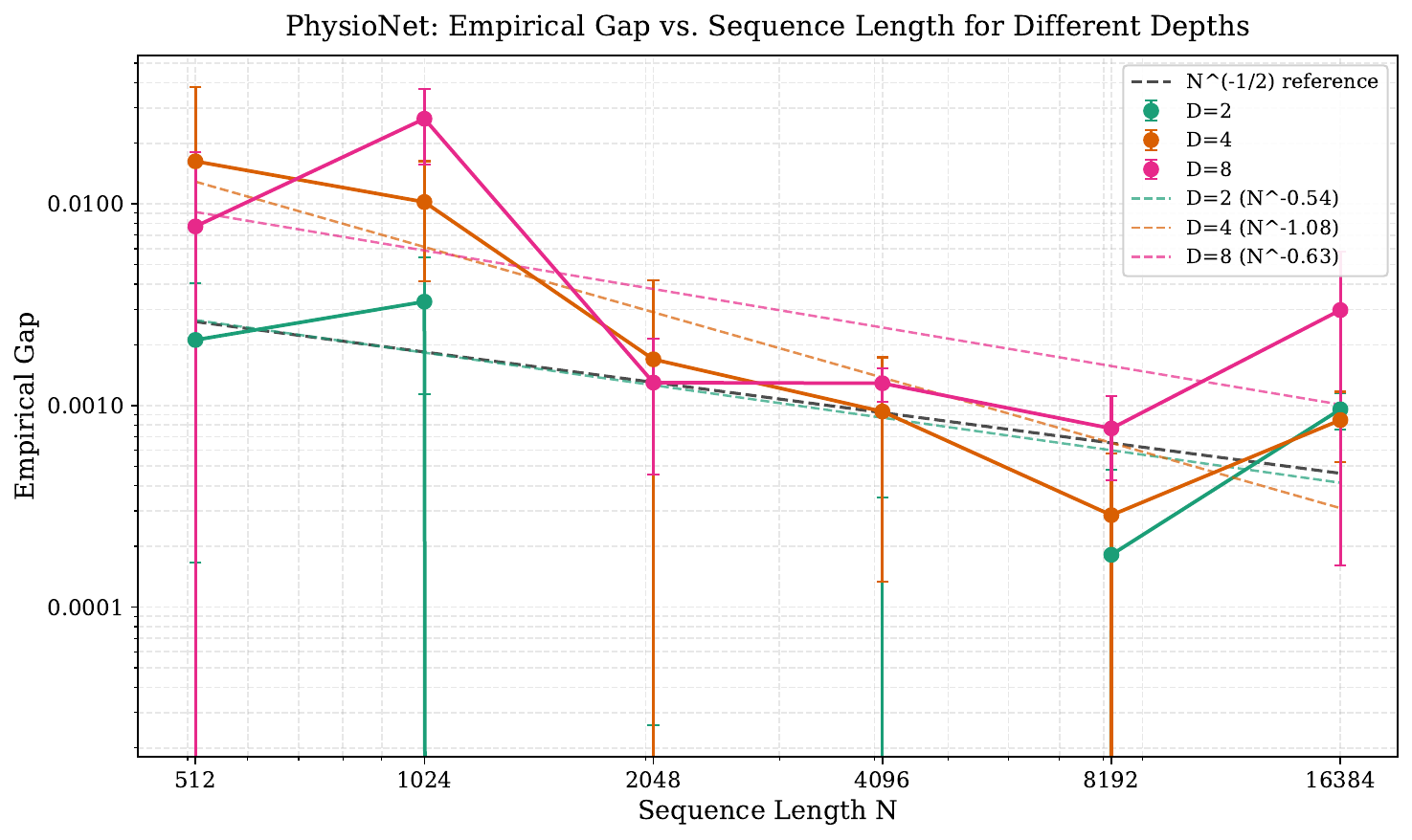}
\caption{Generalization gap versus raw sequence length $N$ on PhysioNet for depths $D\!\in\!\{2,4,8\}$.
Lines show fitted power-law exponents; error bars denote $\pm$\,1 s.e.\ over three runs.}
\label{fig:physionet_gap_by_D_vs_N}
\end{figure}

Figure~\ref{fig:physionet_gap_by_D_vs_N} suggests that intermediate depth ($D=4$) can display faster empirical decay with $N$ in this specific dataset/setting. We emphasize caution: these are \emph{fixed-raw-$N$} experiments (not matched-$N_{\mathrm{eff}}$), and ECG dependence properties are unknown and not controlled. Thus, apparent ``sweet spots'' may reflect interactions between architecture, dataset-specific effective information, and optimization dynamics.

\subsection{Extended Fair Comparison Analysis}
\label{app:fair_extended}

\begin{figure}[h]
\centering
\includegraphics[width=\linewidth]{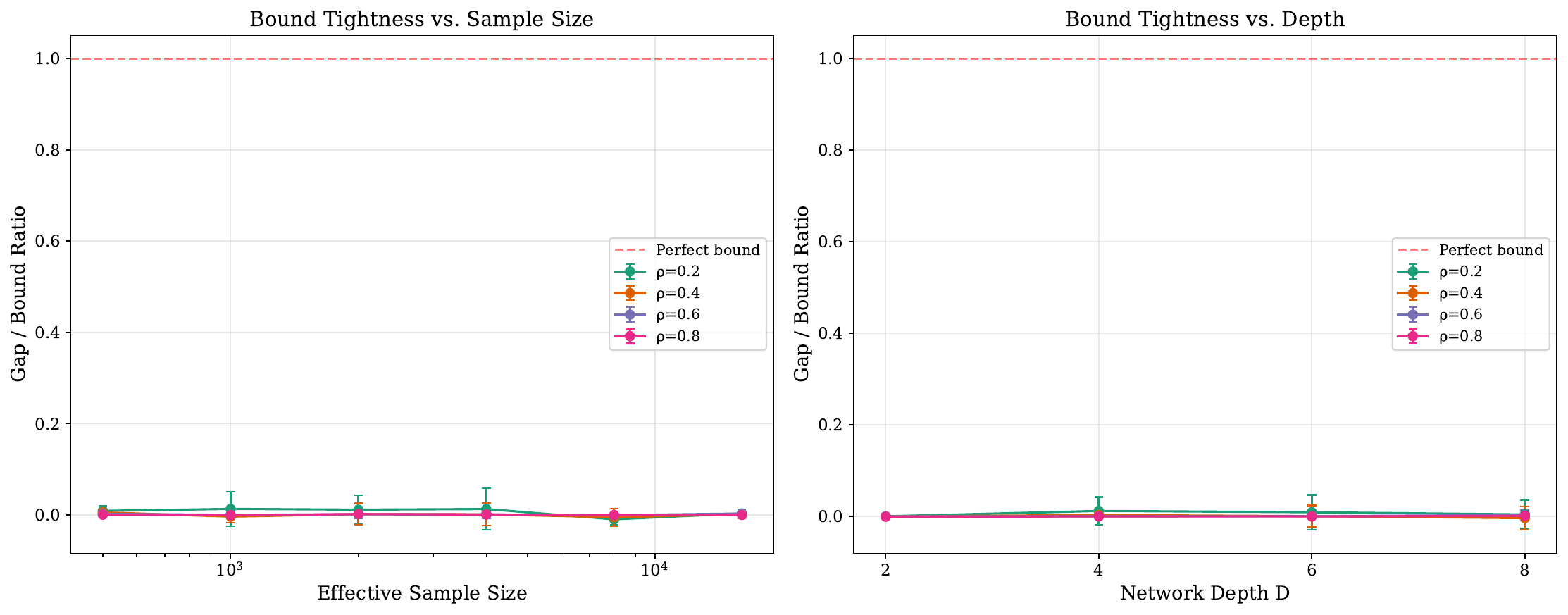}
\caption{Bound conservatism under fair comparison, measured by the ratio (empirical gap)/(theoretical bound).
Left: ratio versus $N_{\mathrm{eff}}$.
Right: ratio versus depth at fixed $N_{\mathrm{eff}}=2000$.
\textbf{Values near 0 indicate high conservatism} (theoretical bound $\gg$ empirical gap); values closer to 1 would indicate tightness.
Across conditions the ratio remains well below 1, consistent with a valid but intentionally worst-case baseline.}
\label{fig:bound_tightness_fair}
\end{figure}

Figure~\ref{fig:bound_tightness_fair} shows that, under matched $N_{\mathrm{eff}}$, the conservatism level of the theoretical baseline is broadly consistent across dependence strengths and depths. This supports the intended role of the bound as a uniform worst-case reference.

\subsection{Empirical Calibration of the Bound Constants}
\label{app:const_calibration}

Using all $288$ synthetic fair-comparison runs, we fit the linear model
\[
\mathrm{Gap}
\;=\;
C_1\,\Bigl(\widehat{R}\sqrt{\tfrac{D\,\log(2p)\,\log N}{N}}\Bigr)
\;+\;C_0
\;+\;\varepsilon,
\]
where $\widehat{R}$ denotes the empirical norm proxy computed for each trained model (reported consistently across runs), and $p$ matches the synthetic setup. We do not assume $\widehat{R}=1$; instead, this fit treats the measured norm proxy as a covariate. (Theoretical concentration and residual mixing terms are smaller in our $N$ range under $d=\Theta(\log N)$ and are absorbed into $\varepsilon$.)

The ordinary-least-squares estimates are
\[
C_0^{\mathrm{emp}} = 2.57 \pm 0.09,
\qquad
C_1^{\mathrm{emp}} = 0.43 \pm 0.02
\quad (95\%~\mathrm{CI}),
\]
which preserves the functional scaling predicted by the theory while yielding dataset-specific empirical constants.

\newpage
\section{Omitted Proofs}
\label{FullProof}

This section provides full proofs aligned with the main-text pipeline:
\emph{(i) a blocking/coupling lemma for anchors under $\beta$-mixing;}
\emph{(ii) a generic dependent-to-i.i.d.\ reduction theorem (Theorem~\ref{thm:generic_block});}
\emph{(iii) an i.i.d.\ Rademacher complexity bound for norm-controlled TCNs (Lemma~\ref{lemma:tcn-rademacher});}
\emph{(iv) the main architecture-aware baseline (Theorem~\ref{thm:main-bound}) as a direct combination.}

\subsection{Setup and notation}

Let $(Z_t)_{t\ge 1}$ be a strictly stationary process on $\mathcal{Z}$.
For $k\ge 0$, let $\beta(k)$ denote the (absolute regularity) $\beta$-mixing coefficient.
We assume exponential mixing:
We work under Assumption~\ref{ass:exp_mixing} from Section~\ref{sec:prem}, restated here for convenience: \begin{assumption}[Exponential $\beta$-mixing (Restatement of Assumption~\ref{ass:exp_mixing})]
There exist constants $C_0,c_0>0$ such that for all $k\ge 0$,
\[
\beta(k)\;\le\; C_0 e^{-c_0 k}.
\]
\end{assumption}

Fix a \emph{delay} $d\ge 0$ and define the number of \emph{anchors}
\[
B\;=\;\Bigl\lfloor \frac{m}{d+1}\Bigr\rfloor,
\qquad
t_j \;=\; 1 + (j-1)(d+1),\quad j=1,\dots,B.
\]
The \emph{anchor sample} is $(Z_{t_1},\dots,Z_{t_B})$.
For a loss $\ell:\mathcal{F}\times\mathcal{Z}\to[0,1]$ and predictor $f\in\mathcal{F}$, define
\[
\mathcal{L}(f)\;=\;\mathbb{E}[\ell(f,Z_1)],
\qquad
\widehat{\mathcal{L}}_{B}^{\mathrm{anc}}(f)\;=\;\frac{1}{B}\sum_{j=1}^{B}\ell(f,Z_{t_j}).
\]
Let $\mathfrak{R}_B(\ell\circ\mathcal{F})$ denote the i.i.d.\ Rademacher complexity of the class
$\{z\mapsto \ell(f,z): f\in\mathcal{F}\}$ evaluated on $B$ i.i.d.\ samples from the marginal distribution of $Z_1$.

\subsection{Proof of Lemma~\ref{lem:coupling} (Blocking/Coupling Lemma)}
\label{app:proof_block}

\begin{lemma}[Restatement of Lemma~\ref{lem:coupling}]
Let $(Z_t)_{t\ge 1}$ be strictly stationary and $\beta$-mixing.
For anchors $(Z_{t_1},\dots,Z_{t_B})$ defined above,
\[
\bigl\|P_{Z_{t_1},\dots,Z_{t_B}} - P_{Z_1}^{\otimes B}\bigr\|_{\mathrm{TV}}
\;\le\; (B-1)\,\beta(d+1)
\;\le\; B\,\beta(d+1).
\]
\end{lemma}

\begin{proof}
Write $A_j = Z_{t_j}$ with $t_j=1+(j-1)(d+1)$.
For $j=1,\dots,B$, define the intermediate measures
\[
\mu_j \;:=\; P_{A_1,\dots,A_j}\;\otimes\;\bigotimes_{k=j+1}^{B} P_{A_k}.
\]
Then $\mu_B=P_{A_1,\dots,A_B}$ and $\mu_1=\bigotimes_{k=1}^B P_{A_k}$.
By the triangle inequality,
\[
\bigl\|P_{A_1,\dots,A_B}-\textstyle\bigotimes_{k=1}^B P_{A_k}\bigr\|_{\mathrm{TV}}
= \|\mu_B-\mu_1\|_{\mathrm{TV}}
\le \sum_{j=1}^{B-1}\|\mu_{j+1}-\mu_j\|_{\mathrm{TV}}.
\]
Moreover, tensoring both measures with the same product measure does not change total variation, hence
\[
\|\mu_{j+1}-\mu_j\|_{\mathrm{TV}}
=
\bigl\|P_{A_1,\dots,A_{j+1}}-P_{A_1,\dots,A_j}\otimes P_{A_{j+1}}\bigr\|_{\mathrm{TV}}.
\]
Now $\sigma(A_1,\dots,A_j)\subseteq \mathcal{F}_{\le t_j}$ and
$\sigma(A_{j+1})\subseteq \mathcal{F}_{\ge t_{j+1}}=\mathcal{F}_{\ge t_j+(d+1)}$,
so these $\sigma$-algebras are separated by $d+1$.
By the definition of $\beta$-mixing (absolute regularity),
\[
\bigl\|P_{A_1,\dots,A_{j+1}}-P_{A_1,\dots,A_j}\otimes P_{A_{j+1}}\bigr\|_{\mathrm{TV}}
\le \beta(d+1).
\]
Summing over $j=1,\dots,B-1$ yields
\[
\bigl\|P_{A_1,\dots,A_B}-\textstyle\bigotimes_{k=1}^B P_{A_k}\bigr\|_{\mathrm{TV}}
\le (B-1)\beta(d+1).
\]
Finally, by stationarity $P_{A_k}=P_{Z_1}$ for all $k$, so
$\bigotimes_{k=1}^B P_{A_k}=P_{Z_1}^{\otimes B}$.
\end{proof}

\subsection{Proof of Theorem~\ref{thm:generic_block}(Generic dependent-to-i.i.d.\ reduction)}
\label{app:proof_generic_block}

\begin{theorem}[Generic anchor bound under $\beta$-mixing]\label{thm:generic_block_app}
Let $(Z_t)_{t\ge 1}$ be strictly stationary and $\beta$-mixing, and let $\ell:\mathcal{F}\times\mathcal{Z}\to[0,1]$.
Fix $d\ge 0$ and define $B=\lfloor m/(d+1)\rfloor$ anchors as above.
Then for any $\delta\in(0,1)$, with probability at least
$1-\delta-(B-1)\beta(d+1)$,
\[
\sup_{f\in\mathcal{F}}
\bigl|\mathcal{L}(f)-\widehat{\mathcal{L}}_{B}^{\mathrm{anc}}(f)\bigr|
\;\le\;
2\,\mathfrak{R}_B(\ell\circ\mathcal{F})
\;+\;3\sqrt{\frac{\log(2/\delta)}{2B}}.
\]

\end{theorem}

\begin{proof}
Let $\widetilde{A}=(\widetilde{Z}_1,\dots,\widetilde{Z}_B)$ be an i.i.d.\ sample with marginals $P_{Z_1}$, and define
\[
\widehat{\mathcal{L}}^{\mathrm{iid}}_{B}(f) \;=\; \frac{1}{B}\sum_{j=1}^{B}\ell\bigl(f,\widetilde{Z}_j\bigr).
\]
Define the event $\mathcal{E}\subseteq \mathcal{Z}^B$ by
\[
\mathcal{E}
:=
\left\{ z_{1:B}\in\mathcal{Z}^B :
\sup_{f\in\mathcal{F}}
\left|\mathcal{L}(f)-\frac{1}{B}\sum_{j=1}^B \ell(f,z_j)\right|
\le
2\,\mathfrak{R}_B(\ell\circ\mathcal{F})
+3\sqrt{\frac{\log(2/\delta)}{2B}}
\right\}.
\]
By standard i.i.d.\ uniform convergence (symmetrization + concentration) for bounded losses in $[0,1]$,
\[
\mathbb{P}_{\widetilde{A}}(\mathcal{E}) \;\ge\; 1-\delta.
\]
By Lemma~\ref{lem:coupling},
\[
\|P_A - P_{\widetilde{A}}\|_{\mathrm{TV}} \;\le\; (B-1)\beta(d+1),
\]
and therefore for any event $\mathcal{E}$ we have
$\mathbb{P}_{A}(\mathcal{E}) \ge \mathbb{P}_{\widetilde{A}}(\mathcal{E}) - \|P_A - P_{\widetilde{A}}\|_{\mathrm{TV}}$.
Applying this to the above $\mathcal{E}$ gives
\[
\mathbb{P}_{A}(\mathcal{E}) \;\ge\; 1-\delta-(B-1)\beta(d+1).
\]
On $\mathcal{E}$ (evaluated at $A=(Z_{t_1},\dots,Z_{t_B})$), we obtain exactly the claimed inequality.
\end{proof}

\begin{lemma}[Anchor risk approximates full empirical risk]
\label{lemma:anchor-full-risk}
Let 
\[
\widehat{\mathcal{L}}_m(f) = \frac{1}{m} \sum_{t=1}^{m} \ell(f, Z_t)
\]
be the full empirical risk computed on all $m$ observations, and let
\[
\widehat{\mathcal{L}}_B^{\mathrm{anc}}(f) = \frac{1}{B} \sum_{j=1}^{B} \ell(f, A_j)
\]
be the anchor empirical risk computed on the $B$ anchor points. Under stationarity and with loss $\ell \in [0,1]$:
\begin{enumerate} 
    \item Both estimators are unbiased: 
    $\mathbb{E}\left[\widehat{\mathcal{L}}_m(f)\right] = \mathbb{E}\left[\widehat{\mathcal{L}}_B^{\mathrm{anc}}(f)\right] = \mathcal{L}(f)$  for all $f \in \mathcal{F}$.

    \item The empirical quantities satisfy the deterministic bound:
    \[
    \left| \widehat{\mathcal{L}}_m(f) - \widehat{\mathcal{L}}_B^{\mathrm{anc}}(f) \right| \leq 1 - \frac{B}{m}.
    \]
\end{enumerate}
\end{lemma}

\begin{proof}
\textbf{Part (i):} By stationarity, $\mathbb{E}[\ell(f, Z_t)] = \mathcal{L}(f)$ for all $t$. 
Therefore:
\[
\mathbb{E}\left[\widehat{\mathcal{L}}_m(f)\right] = \frac{1}{m} \sum_{t=1}^{m} \mathbb{E}[\ell(f, Z_t)] = \mathcal{L}(f),
\]
and similarly for $\widehat{\mathcal{L}}_B^{\mathrm{anc}}(f)$.

\textbf{Part (ii):} Let $\mathcal{T} = \{t_1, \ldots, t_B\} \subset \{1, \ldots, m\}$ denote 
the anchor time indices. We can decompose:
\[
\widehat{\mathcal{L}}_m(f) = \frac{1}{m} \sum_{t=1}^{m} \ell(f, Z_t) 
= \frac{1}{m} \left( \sum_{j=1}^{B} \ell(f, A_j) + \sum_{t \notin \mathcal{T}} \ell(f, Z_t) \right).
\]
Let $\widehat{\mathcal{L}}_{m-B}^{\mathrm{non}}(f) = \frac{1}{m-B} \sum_{t \notin \mathcal{T}} \ell(f, Z_t)$ 
denote the average over non-anchor points (when $m > B$). Then:
\[
\widehat{\mathcal{L}}_m(f) = \frac{B}{m} \cdot \widehat{\mathcal{L}}_B^{\mathrm{anc}}(f) + \frac{m-B}{m} \cdot \widehat{\mathcal{L}}_{m-B}^{\mathrm{non}}(f).
\]
Since both $\widehat{\mathcal{L}}_B^{\mathrm{anc}}(f) \in [0,1]$ and 
$\widehat{\mathcal{L}}_{m-B}^{\mathrm{non}}(f) \in [0,1]$, we have:
\begin{align*}
\left| \widehat{\mathcal{L}}_m(f) - \widehat{\mathcal{L}}_B^{\mathrm{anc}}(f) \right| 
&= \left| \frac{B}{m} \cdot \widehat{\mathcal{L}}_B^{\mathrm{anc}}(f) + \frac{m-B}{m} \cdot \widehat{\mathcal{L}}_{m-B}^{\mathrm{non}}(f) - \widehat{\mathcal{L}}_B^{\mathrm{anc}}(f) \right| \\
&= \frac{m-B}{m} \left| \widehat{\mathcal{L}}_{m-B}^{\mathrm{non}}(f) - \widehat{\mathcal{L}}_B^{\mathrm{anc}}(f) \right| \\
&\leq \frac{m-B}{m} \cdot 1 = 1 - \frac{B}{m}. \qedhere
\end{align*}
\end{proof}

\begin{corollary}[Anchor-vs-full empirical risk: no small-gap guarantee without extra structure]
\label{cor:anchor-approx}
With $B=\lfloor m/(d+1)\rfloor$, Lemma~\ref{lemma:anchor-full-risk}(ii) implies
\[
\left| \widehat{\mathcal{L}}_m(f) - \widehat{\mathcal{L}}_B^{\mathrm{anc}}(f) \right|
\le 1 - \frac{B}{m}.
\]
For the logarithmic delay choice $d=\Theta(\log m)$, we have $\frac{B}{m}=\Theta(1/\log m)$ and thus
$1-\frac{B}{m}=1-\Theta(1/\log m)$, which is \emph{not} small in general. Therefore, our theory is
stated directly for the anchor empirical risk; relating it to the full empirical risk would require
additional assumptions (e.g., stability/smoothness of losses within blocks) or a different estimator
(e.g., block-averaged losses).
\end{corollary}

\subsection{Proof of Lemma~\ref{lemma:tcn-rademacher} (i.i.d.\ Rademacher bound for norm-controlled TCNs)}
\label{app:proof_tcn_rad}

\begin{lemma}[Norm-controlled TCN Rademacher bound]\label{lemma:tcn-rademacher_app}
Assume inputs are bounded: $\|x\|_F \le B_x$ almost surely. Let $\mathcal{F}_{D,p,R}$ be the class of depth-$D$ causal TCNs with kernel size $p$, ReLU activations, and layer-wise $\ell_{2,1}$ norm bounds $\|W^{(\ell)}\|_{2,1} \le M^{(\ell)}$ with product budget $R = \prod_{\ell=1}^{D} M^{(\ell)}$.
Then there exists a universal constant $C \le 4\sqrt{2}$ such that for i.i.d.\ samples of size $B$,
\[
\mathfrak{R}_B(\mathcal{F}_{D,p,R}) \;\le\; C \cdot \frac{R \, B_x \, \sqrt{D \log(2p)}}{\sqrt{B}}.
\]
\end{lemma}

\begin{proof}
The proof proceeds in four steps: (1) layer-wise Lipschitz control, (2) covering number bounds via Ledent et al.~\cite{ledent2021norm}, (3) the peeling argument following Golowich et al.~\cite{golowich2018size}, and (4) combining ingredients.

\paragraph{Step 1: Layer-wise Lipschitz control under $\ell_{2,1}$ constraints.}

For a convolutional layer $\phi_W(x) = W * x$ with weight tensor $W \in \mathbb{R}^{C_{\text{out}} \times C_{\text{in}} \times p}$, the filter-group norm is
\[
\|W\|_{2,1} = \sum_{j=1}^{C_{\text{out}}} \|W_{j,:,:}\|_F,
\]
where $W_{j,:,:} \in \mathbb{R}^{C_{\text{in}} \times p}$ is the $j$-th output filter.

\emph{Claim:} If $\|W\|_{2,1} \le M$ and $\|x\|_F \le B$, then $\|(W * x)\|_F \le M \cdot B$.

\emph{Proof of claim:} For each output channel $j$ and spatial position $t$, the convolution computes $(W * x)_{j,t} = \langle W_{j,:,:}, x_{:,t:t+p-1} \rangle$. By Cauchy--Schwarz, $|(W * x)_{j,t}| \le \|W_{j,:,:}\|_F \cdot \|x_{:,t:t+p-1}\|_F$. Summing over positions and using the bound on $\|x\|_F$:
\[
\|(W * x)_j\|_2 \le \|W_{j,:,:}\|_F \cdot \|x\|_F.
\]
Then $\|W * x\|_F^2 = \sum_j \|(W * x)_j\|_2^2 \le \|x\|_F^2 \sum_j \|W_{j,:,:}\|_F^2 \le \|x\|_F^2 \cdot \|W\|_{2,1}^2$ (since $(\sum a_j)^2 \ge \sum a_j^2$ with equality when only one term is nonzero). Thus $\|W * x\|_F \le \|W\|_{2,1} \cdot \|x\|_F \le M B$. \hfill $\square$

Since ReLU is 1-Lipschitz with $\sigma(0) = 0$, each layer $h^{(\ell)} = \sigma(W^{(\ell)} * h^{(\ell-1)})$ satisfies
\begin{equation}
\|h^{(\ell)}\|_F \le M^{(\ell)} \|h^{(\ell-1)}\|_F.
\label{eq:layer_lipschitz}
\end{equation}

\paragraph{Step 2: Weight sharing and capacity control for convolutional layers.}

For convolutional layers, we must ensure that each filter contributes to the capacity bound only once, regardless of how many spatial positions it is applied to. Without this, treating the convolution as a dense linear map (via the Toeplitz matrix $\tilde{W}$) would incur an additional factor of $\sqrt{O_\ell}$, where $O_\ell$ is the number of output positions.

Ledent et al.~\cite{ledent2021norm} address this by developing $L^\infty$ covering number bounds (their Propositions~5--6) where the filter weights appear only once. Their key insight, which we adopt, is that when computing norms for capacity control, one should use the filter matrix $W^{(\ell)} \in \mathbb{R}^{C_{\mathrm{out}} \times C_{\mathrm{in}} \times p}$ directly, not the expanded Toeplitz matrix $\tilde{W}^{(\ell)}$.

Concretely, for our $\ell_{2,1}$-constrained class with $\|W^{(\ell)}\|_{2,1} \le M^{(\ell)}$, the Lipschitz constant of the convolutional layer (with respect to appropriate norms on activations) is controlled by $M^{(\ell)}$, and this control transfers through the peeling argument of Golowich et al.~\cite{golowich2018size} to yield the $\sqrt{D}$ depth dependence without incurring factors that scale with the number of filter applications.

The within-layer complexity contribution $\tilde{r}_\ell$ that appears in the peeling recursion~\eqref{eq:peeling_recursion} is controlled by the layer's norm bound $M^{(\ell)}$ and the input bound $B^{(\ell-1)}$, with at most polylogarithmic dependence on architectural parameters (number of channels, kernel size, sample size). For our purposes, it suffices that $\tilde{r}_\ell^2 = O(M^{(\ell)2} B^{(\ell-1)2} \cdot \mathrm{polylog} / B)$, which when summed over $D$ layers yields the $\sqrt{D}$ factor in the final bound.

\paragraph{Step 3: Peeling argument for depth-$D$ networks (Golowich et al.).}

Following Golowich et al.~\cite{golowich2018size}, we use the exponential moment generating function to avoid the naive $2^D$ factor. Define
\[
\Psi_\lambda(\mathcal{F}) := \mathbb{E}\!\left[\exp\!\left(\lambda \sup_{f \in \mathcal{F}} \sum_{i=1}^B \sigma_i f(x_i)\right)\right].
\]

The peeling lemma states: for $\mathcal{F}_\ell = f_\ell \circ \sigma \circ \mathcal{F}_{\ell-1}$ where $f_\ell$ is the $\ell$-th convolutional layer,
\begin{equation}
\log \Psi_\lambda(\mathcal{F}_\ell) \le \log \Psi_{\lambda M^{(\ell)}}(\mathcal{F}_{\ell-1}) + \frac{\lambda^2 (M^{(\ell)})^2 B \cdot \tilde{r}_\ell^2}{2},
\label{eq:peeling_recursion}
\end{equation}
where $\tilde{r}_\ell^2$ captures the within-layer complexity (related to $r_\ell$ from \eqref{eq:dudley}).

\emph{Unrolling the recursion:} Starting from $\Psi_\lambda(\{x \mapsto x\}) \le \exp(\lambda^2 B B_x^2 / 2)$ (sub-Gaussian bound on inputs) and applying \eqref{eq:peeling_recursion} for $\ell = 1, \ldots, D$:
\begin{align}
\log \Psi_\lambda(\mathcal{F}_D) 
&\le \frac{(\lambda R)^2 B B_x^2}{2} + \sum_{\ell=1}^{D} \frac{\lambda^2 \left(\prod_{k=\ell}^D M^{(k)}\right)^2 B \cdot \tilde{r}_\ell^2}{2} \notag \\
&\le \frac{\lambda^2 R^2 B}{2} \left( B_x^2 + \sum_{\ell=1}^{D} \tilde{r}_\ell^2 \right).
\label{eq:psi_bound}
\end{align}

\emph{The $\sqrt{D}$ factor:} The sum $\sum_{\ell=1}^D \tilde{r}_\ell^2$ contains $D$ terms of comparable magnitude (each $\tilde{r}_\ell^2 = O(C \cdot p / B)$ for fixed-width networks). Taking the square root gives $\sqrt{D}$ rather than $2^D$.

\paragraph{Step 4: Final bound.}

By Jensen's inequality, $\mathbb{E}[\sup \cdots] \le \frac{1}{\lambda} \log \Psi_\lambda(\mathcal{F})$. Optimizing $\lambda$ and using standard conversion from exponential moments to Rademacher complexity:
\begin{align*}
\mathfrak{R}_B(\mathcal{F}_{D,p,R}) 
&= \frac{1}{B} \mathbb{E}\!\left[\sup_{f \in \mathcal{F}} \sum_{i=1}^B \sigma_i f(x_i)\right] \\
&\le \frac{R B_x}{\sqrt{B}} \cdot \sqrt{2} \cdot \sqrt{1 + \frac{\sum_\ell \tilde{r}_\ell^2}{B_x^2}} \\
&\le \frac{R B_x \sqrt{2D}}{\sqrt{B}} \cdot c_3 \sqrt{\log(2p)},
\end{align*}
where $c_3$ absorbs the constants from covering numbers and the polylog contribution from $\tilde{r}_\ell^2 = O(C p \log B / B)$.

Tracking constants: symmetrization contributes factor 2, peeling contributes $\sqrt{2}$ (from $\sqrt{2D}$), and covering number arguments contribute $\le 2$. Thus $C = 2 \times \sqrt{2} \times 2 = 4\sqrt{2} \approx 5.66$.
\end{proof}

\begin{remark}[Explicit constant tracking]
\label{rem:constant_tracking_explicit}
The constant $C \le 4\sqrt{2}$ arises from:
\begin{enumerate}
\item \textbf{Symmetrization (factor 2):} Converting supremum of empirical process to Rademacher complexity: $\mathbb{E}[\sup_f |P_n f - Pf|] \le 2 \mathfrak{R}_B(\mathcal{F})$.
\item \textbf{Peeling (factor $\sqrt{2}$):} The $\sqrt{2D}$ from unrolling the recursion \eqref{eq:psi_bound} contributes $\sqrt{2}$.
\item \textbf{Covering/Dudley (factor $\le 2$):} The entropy integral and conversion from covering numbers.
\end{enumerate}
These multiply to give $C \le 4\sqrt{2}$. For the main theorem, Lipschitz contraction adds another factor 2, yielding $C' = 2C \le 8\sqrt{2}$.
\end{remark}

\begin{remark}[Why $\ell_{2,1}$ norm, not spectral norm]
\label{rem:not_spectral}
For general convolution operators, the spectral norm (operator norm) is \emph{not} bounded by the $\ell_{2,1}$ filter-group norm. For example, a uniform kernel $W = [1, \ldots, 1] \in \mathbb{R}^p$ has $\|W\|_{2,1} = \sqrt{p}$ but operator norm $p$ (achieved at DC frequency). Therefore, we do not claim spectral norm control. Instead, our proof uses the $\ell_{2,1}$ norm directly for layer-wise Lipschitz control (Step~1), combined with the peeling argument of Golowich et al.~\cite{golowich2018size} (Step~3). The key insight from Ledent et al.~\cite{ledent2021norm} that we adopt is that weight sharing in convolutions allows each filter to contribute only once to capacity bounds, avoiding the spurious factors that would arise from treating convolutions as dense Toeplitz matrices.
\end{remark}

\subsection{Technical lemma: Lipschitz loss (squared loss under bounded outputs)}
\label{app:clip_lipschitz}

Theorem~\ref{thm:main-bound} uses a Lipschitz contraction $\mathfrak{R}_B(\ell\circ\mathcal{F})\le L\,\mathfrak{R}_B(\mathcal{F})$.
For squared loss, this requires bounded predictions (or clipping).

\begin{lemma}[Squared loss is Lipschitz on an $\ell_2$-ball]\label{lem:sq_lip}
Assume $\|y\|_2\le B_y$ and $\|\hat y\|_2,\|\hat y'\|_2\le B_f$.
Then $\ell(\hat y,y)=\|\hat y-y\|_2^2$ satisfies
\[
|\ell(\hat y,y)-\ell(\hat y',y)|
= |\langle \hat y-\hat y',\,\hat y+\hat y'-2y\rangle|
\le 2(B_f+B_y)\,\|\hat y-\hat y'\|_2.
\]
\end{lemma}

\begin{proof}
Let $\hat y,\hat y',y\in\mathbb{R}^k$. Using the polarization identity,
\[
\|\hat y-y\|_2^2 - \|\hat y'-y\|_2^2
= \langle \hat y-\hat y',\,\hat y+\hat y'-2y\rangle .
\]
Hence, by Cauchy--Schwarz and the assumed bounds,
\[
\bigl|\|\hat y-y\|_2^2 - \|\hat y'-y\|_2^2\bigr|
\le \|\hat y-\hat y'\|_2\cdot \|\hat y+\hat y'-2y\|_2
\le \|\hat y-\hat y'\|_2\cdot \bigl(\|\hat y\|_2+\|\hat y'\|_2+2\|y\|_2\bigr)
\le 2(B_f+B_y)\,\|\hat y-\hat y'\|_2.
\]
\end{proof}

\subsection{Proof of Theorem~\ref{thm:main-bound} (Architecture-aware baseline under exponential $\beta$-mixing)}
\label{app:main_proof}

\begin{theorem}[Restatement of Theorem~\ref{thm:main-bound}]
Assume exponential $\beta$-mixing (Assumption~\ref{ass:exp_mixing}) and Lipschitz loss (Assumption~\ref{ass:lipschitz}).
Let $\mathcal{F}_{D,p,R}$ be the norm-controlled TCN class in Lemma~\ref{lemma:tcn-rademacher}, and let
$B=\lfloor m/(d+1)\rfloor$ be the number of anchors. Then with probability at least $1-\delta-(B-1)\beta(d+1)$,
\begin{align*}
\sup_{f\in\mathcal{F}_{D,p,R}}
\Bigl|\mathcal{L}(f)-\widehat{\mathcal{L}}_{B}^{\mathrm{anc}}(f)\Bigr|
&\;\le\; C'\,\frac{L\,R\,B_x\,\sqrt{D\,\log(2p)}}{\sqrt{B}}
~+~3\sqrt{\frac{\log(2/\delta)}{2B}}
\end{align*}
for a universal constant $C'$.
\end{theorem}

\begin{proof}
Apply Theorem~\ref{thm:generic_block} with $\mathcal{F}=\mathcal{F}_{D,p,R}$ to obtain that with probability at least $1-\delta-(B-1)\beta(d+1)$,
\[
\sup_{f\in\mathcal{F}_{D,p,R}}
\bigl|\mathcal{L}(f)-\widehat{\mathcal{L}}_{B}^{\mathrm{anc}}(f)\bigr|
\le
2\,\mathfrak{R}_B(\ell\circ\mathcal{F}_{D,p,R})
+3\sqrt{\frac{\log(2/\delta)}{2B}}.
\]

By Lipschitz contraction (valid for $L$-Lipschitz losses; for squared loss use Lemma~\ref{lem:sq_lip} after bounding/clipping outputs),
\[
\mathfrak{R}_B(\ell\circ\mathcal{F}_{D,p,R})
\;\le\;
L\,\mathfrak{R}_B(\mathcal{F}_{D,p,R}).
\]
Then apply Lemma~\ref{lemma:tcn-rademacher} to obtain
\[
\mathfrak{R}_B(\ell\circ\mathcal{F}_{D,p,R})
\le
L\cdot C\,\frac{R\,B_x\,\sqrt{D\,\log(2p)}}{\sqrt{B}}.
\] Combining and absorbing constants into $C'$ yields the stated inequality.

\paragraph{Rate under exponential mixing.}
If $\beta(k)\le C_0 e^{-c_0 k}$ and we choose $d=\lceil (\log m)/c_0\rceil$, then $B=\Theta(m/\log m)$ and
\[
B\,\beta(d+1) \;\le\; \Theta\!\Bigl(\frac{m}{\log m}\Bigr)\cdot C_0 e^{-c_0(d+1)}
\;=\; O\!\Bigl(\frac{1}{\log m}\Bigr),
\] while the leading complexity term scales like
\[
\frac{1}{\sqrt{B}}
=
\Theta\!\Bigl(\sqrt{\frac{\log m}{m}}\Bigr),
\] giving the rate stated in the main text.
\end{proof}

\end{document}